\documentclass[letterpaper, 10 pt, journal, twoside]{IEEEtran}
\usepackage{amsmath}
\usepackage{amsfonts}
\usepackage{algorithmic}
\usepackage{algorithm}
\usepackage{array}
\usepackage[caption=false,font=normalsize,labelfont=sf,textfont=sf]{subfig}
\usepackage{textcomp}
\usepackage{stfloats}
\usepackage{url}
\usepackage{verbatim}
\usepackage{graphicx}
\usepackage{cite}
\usepackage{bbding}
\usepackage{siunitx}
\usepackage{bm}
\usepackage{color}
\usepackage{xcolor}
\newcommand{\red}[1]{\textcolor{black}{#1}}
\newcommand{\rred}[1]{\textcolor{black}{#1}}

\begin{document}

\title{\red{A Versatile Neural Network Configuration }Space Planning and Control Strategy for Modular Soft Robot Arms}
\author{Zixi Chen, 
Qinghua Guan, 
Josie Hughes,~\IEEEmembership{Member,~IEEE}, 
Arianna Menciassi,~\IEEEmembership{Fellow,~IEEE}, 
and Cesare Stefanini,~\IEEEmembership{Member,~IEEE}
\thanks{This work was supported by the European Union by the Next Generation EU project ECS00000017 ‘Ecosistema dell’Innovazione’ Tuscany Health Ecosystem (THE, PNRR, Spoke 9: Robotics and Automation for Health.) 
Z. Chen and Q. Guan contributed equally to this work.
\emph{*Corresponding authors: Zixi Chen}} 
\thanks{This work has been submitted to the IEEE for possible publication. Copyright may be transferred without notice, after which this version may no longer be accessible.} 
\thanks{Zixi Chen, Arianna Menciassi, and Cesare Stefanini are with the Biorobotics Institute and the Department of Excellence in Robotics and AI, Scuola Superiore Sant’Anna, 56127 Pisa, Italy (e-mail: zixi.chen@santannapisa.it; arianna.menciassi@santannapisa.it; cesare.stefanini@santannapisa.it).}
\thanks{Qinghua Guan and Josie Hughes are with CREATE Lab, EPFL,
Lausanne, Switzerland (e-mail: qinghua.guan@epfl.ch; josie.hughes@epfl.ch).}
\thanks{Digital Object Identifier (DOI): see top of this page.}
}

\markboth{IEEE TRANSACTION ON ROBOTICS. PREPRINT VERSION. ACCEPTED June 2025} 
{Chen \MakeLowercase{\textit{et al.}}: A Versatile Neural Network Configuration Space Planning and Control Strategy for Modular Soft Robot Arms} 

\maketitle
\begin{abstract}
\rred{Modular soft robot arms (MSRAs) are composed of multiple modules connected in a sequence, and they can bend at different angles in various directions. This capability allows MSRAs to perform more intricate tasks than single-module robots. However, the modular structure also induces challenges in accurate planning and control. Nonlinearity and hysteresis complicate the physical model, while the modular structure and increased DOFs further lead to cumulative errors along the sequence. To address these challenges, we propose a versatile configuration space planning and control strategy for MSRAs, named $S2C2A$ (State to Configuration to Action). Our approach formulates an optimization problem, $S2C$ (State to Configuration planning), which integrates various loss functions and a forward model based on biLSTM to generate configuration trajectories based on target states. A configuration controller $C2A$ (Configuration to Action control) based on biLSTM is implemented to follow the planned configuration trajectories, leveraging only inaccurate internal sensing feedback. We validate our strategy using a cable-driven MSRA, demonstrating its ability to perform diverse offline tasks such as position and orientation control and obstacle avoidance. Furthermore, our strategy endows MSRA with online interaction capability with targets and obstacles. Future work focuses on addressing MSRA challenges, such as more accurate physical models.}
\end{abstract}

\section{Introduction}
\label{sec1}
\IEEEPARstart{M}{ost} soft robots are made of elastomeric materials \cite{14MC} or built using deformable structures \cite{23QG}.
These materials and structures contribute to the lower stiffness and enhanced safety of soft robots compared to their traditional rigid counterparts. 
Due to their safety, they have been leveraged for various applications, including medical scenarios \cite{22ZT}. 
Also, resulting from their compliance, soft robots exhibit high degrees of freedom (DOFs) and, hence, are able to perform environment exploration tasks\cite{14AMc}.
Consequently, soft robots have been successfully deployed in complex working environments, such as the abdomen \cite{14MC} and cluttered pipe environments \cite{14JL}.

Among the various categories of soft robots, modular soft robot arms (MSRA) have demonstrated significant potential, particularly due to their higher DOFs than single-module robots.
\red{Typically, one MSRA is composed of multiple independent modules connected in series, which can be different \cite{16TT} or uniform \cite{24ZCza}.
“Soft continuum manipulators’ \cite{14AMc} or ’continuum robots’ \cite{24KQ} can contain multiple modules and also fall into the MSRA category. 
We use ’modular soft robot arm’ instead of these general and common names to emphasize modularity, which is the focus of this work.
}
The presence of multiple modules increases the degrees of actuation within the same robot length, providing the robots with richer working spaces and more diverse shape patterns \cite{23QG}. 
Due to these unique advantages, MSRAs have been leveraged in challenging tasks such as obstacle avoidance \cite{24KQ} and challenging environment exploration \cite{16HW}.

\begin{figure*}[!ht]
\centering
\includegraphics[width=7.1in]{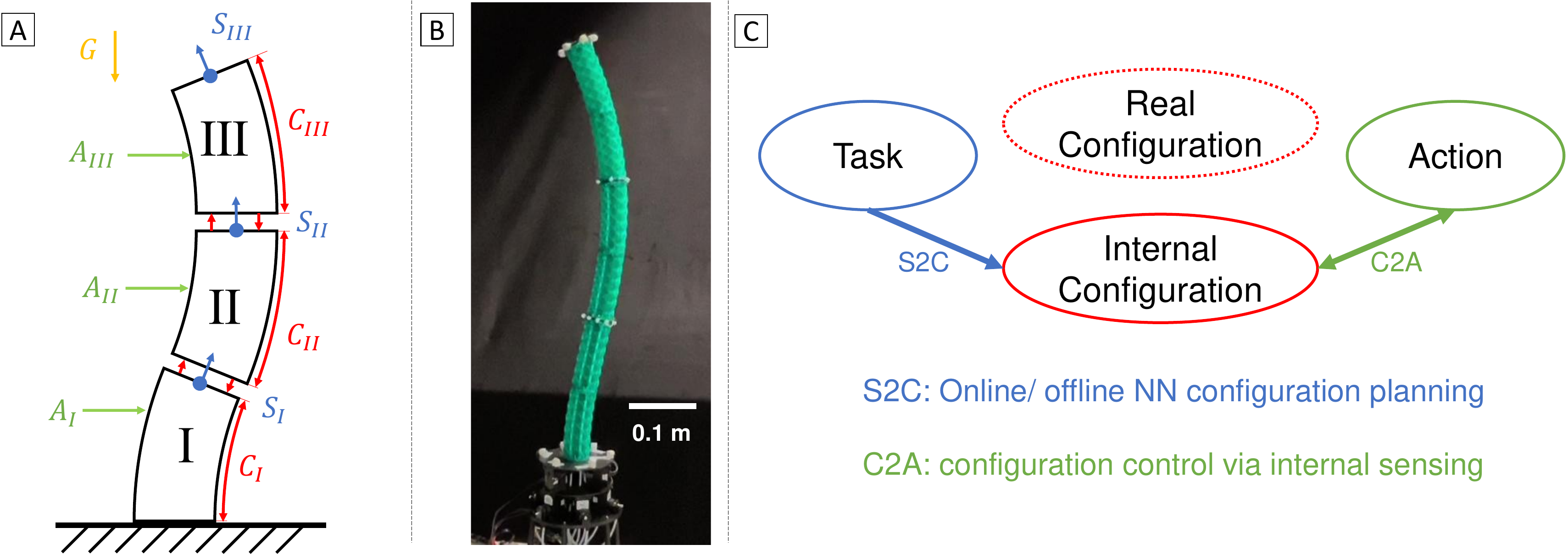}
\caption{
(A) Modular soft robot arm motion diagram. The action of each module $A_*$ (green) will directly affect the configurations $C_*$ of each module (red) and indirectly affect the whole robot system state $S_*$ (blue). The configurations of modules are also affected by gravity and the adjacent modules.
(B) MSRA applied in our experiments can deform in an S-shape, which cannot be achieved by single-module robots.
\red{(C) S2C2A comprises a configuration planning strategy S2C and a configuration controller C2A. 
NN can contribute to the S2C strategy and, based on the various task requests, propose the target configuration online or offline. 
The other NN can serve as a C2A controller, collecting configuration estimation from internal sensors and determining action based on sensor feedback and target configuration from S2C.
This approach can achieve accurate MSRA state control without the knowledge of real configurations.
}}
\label{fig1}
\end{figure*}

Along with the high DOFs, MSRAs also possess some difficulties unique to single-module soft robots, mainly induced by the complicated mappings among the task, configuration, and actuation spaces.
In single-module robot research, most works map directly between the task and actuation space, and the configuration space is ignored \cite{23ZCb}.
\red{However, the module configurations play a significant role in constructing the robot shape in the task space, as shown in Fig. \ref{fig1}(A).}
Various module features can be applied as the configuration, like bending angle and direction \cite{14JL} and module end orientation \cite{24ZCza}.
\red{To measure the MSRA configuration, external sensors like the optical tracking systems \cite{24JL} and internal sensors like inertial measurement unit (IMU) \cite{24GP} and motor encoder \cite{23QG} are leveraged.
However, most internal sensing approaches suffer from low accuracy caused by cable slack and drift \cite{24GP}. 
Targeted and intelligent algorithms are required to cope with sensing inaccuracy.}

Furthermore, the mapping between the task and configuration space induces redundancy in MSRA control. 
Most control targets, like position and orientation, can be managed by multiple configuration choices \cite{24JL}. 
This kind of redundancy endows MSRA with adaptability to various high-level tasks, such as obstacle avoidance. 
Meanwhile, proper algorithms are excepted to determine the configuration choice and fully achieve this potential.
Additionally, modularity also induces difficulty in configuration control.
In single-module robot research, although delay, hysteresis, and nonlinearity have already posed challenges in accurate control, the robot state basically relies on the previous states and actions.
In addition to these challenges, configuration control in MSRA should also consider the influence of the neighboring modules \cite{24ZCza} as shown in Fig. \ref{fig1}(A).
Overall, MSRA control requests an unique configuration planning strategy and a configuration controller.

\begin{table*}[!htb]
\caption{\red{Related works of MSRA}}
\centering
\begin{tabular}{p{0.3in} p{0.5in} p{0.35in} p{0.35in} p{0.6in} p{1.5in} p{2.0in}}
Ref. & Task & Module number & Dimension & Model & Planning & Details \\
\hline
\cite{14AMc} & position & 6 & 2 & PCC & forward model + optimization & confined environment exploration \\
\cite{15AMb} & position & 4 & 2 & pseudo rigid robot model & forward model + optimization & target reaching \\
\cite{24MD} & position & 2 & 3 & PCC & artificial potential field & obstacle avoidance \\
\cite{14JL} & position & 3 & 3 & PCC & inverse model + sampling & simulation, cluttered space exploration \\
\cite{16HW} & position & 6 & 2 & PCC & GMR & learning from demonstration, passing holes \\
\cite{22BM} & position & 3 & 3 & PCC & RRT & obstacle avoidance \\
\hline
\cite{22JL} & position + orientation & 2 & 3 & PCC & forward model + optimization & obstacle avoidance\\
\cite{23QG}  & position + orientation & 3 & 3 & PCC & forward model + optimization & interaction, our previous work\\
\cite{24KQ} & position + orientation & 3 & 3 & PCC & multiple configuration solutions + optimization & obstacle avoidance\\
\cite{24JL} & position + orientation & 3 & 3 & PCC & fuzzy logic controller & close loop for disturbance rejection\\
\hline
\cite{18BO} & shape & 3 & 3 & PCC & user-defined curve + optimization & fitting drawn curve\\
\cite{22ZM} & shape & 2 & 3 & Biarc and Bézier Curve & inverse model & passing tube\\
\hline
Ours & shape & 3 & 3 & NN & NN + optimization & internal sensing\\
\end{tabular}
\label{table1}
\end{table*}

\red{Works related to MSRA planning and control are summarized in Table \ref{table1}. 
Similar to single-module robot control, the end position is the most common control target. 
To cope with the challenging mappings among the task, configuration, and actuation spaces, optimization based on a forward model is an effective solution. 
For instance, a planar MSRA is modeled as Piecewise Constant Curvature (PCC) in \cite{14AMc}, and one optimization approach is employed to plan proper configurations based on task requirements like position control and collision avoidance. 
Similarly, optimization in \cite{15AMb} directly maps from the task space to the actuation space based on the pseudo-rigid robot model. 
In addition to the optimization, sampling based on the inverse model can also contribute to configuration planning, and this strategy can achieve tasks like cluttered space exploration \cite{14JL}.
Rapidly exploring random trees (RRT) samples the configuration space in \cite{22BM} and proposes feasible trajectories.
The Gaussian Mixture Model (GMM) is deployed in \cite{16HW} as the inverse model from task space to actuation space and can generate trajectory learning from the demonstration.}

\red{In addition to the position control task, some unique tasks can only be achieved by MSRAs, such as position and orientation control \cite{23QG} and shape control \cite{18BO}. 
Similar to the position control strategy, forward models like PCC \cite{22JL} and optimization are utilized for configuration planning based on target position and orientation. 
Inverse kinematics is also applied to propose multiple configuration solutions according to the target position and orientation \cite{24KQ}. 
Additionally, the fuzzy logic controller \cite{24JL} directly maps between the task and actuation space and can keep end position/orientation invariant while changing orientation/position. 
Beyond end position and orientation control, the MSRA shape can also be controlled, fully achieving the potential of its structure. 
Optimization \cite{18BO} and inverse model \cite{22ZM} are also leveraged for shape control.}

Considering the inaccurate internal configuration estimation and the challenges of MSRA control, we leverage neural networks (NNs) in this work.
NNs, particularly recurrent neural networks (RNNs), have been extensively leveraged in soft robot modeling and control. 
Due to the nonlinear activation functions, NNs are suitable for handling the nonlinear motions characteristic of soft robots. 
Moreover, the recurrent structures of RNN can be employed to address sequence-related issues such as time delay \cite{18TT} and MSRA module sequence \cite{24ZCza}. 
Several reviews have also highlighted the effectiveness of this data-driven approach in soft robot modeling and control \cite{24ZCa, 23CL}.
\red{In addition, as a data-driven approach, RNN has the ability to map between spaces without explicit physical relationships, like the MSRA task space and the configuration space constructed by the inaccurate sensing feedback, as shown in Fig. \ref{fig1}(C).}
Therefore, we believe that RNNs have the potential to tackle the complexities posed by MSRAs.

This paper aims to propose a versatile MSRA configuration space planning and control strategy named \textit{S2C2A} based on NNs and internal sensing feedback, as illustrated in Fig. \ref{fig1}(C). 
Internal sensors provide rough sensing feedback to represent module configuration.
A configuration planning approach \textit{S2C} via optimization is employed to generate configuration trajectories based on the task requirements online or offline, and we apply a biLSTM network as the MSRA forward model in optimization.
Additionally, a biLSTM configuration controller \textit{C2A} is designed specifically for MSRAs, and the rough internal configuration feedback is employed to perform configuration control tasks.
\red{We validate our approach on the cable-driven MSRA shown in Fig. \ref{fig1}(B) and compare it with our previous work \cite{23QG}, demonstrating its versatility by achieving a variety of tasks.}

\red{The contributions of this paper are as follows:
\begin{enumerate}
\item We utilize space sequence biLSTM serving as the MSRA forward model in the configuration planning optimization problem for online and offline planning.
\item We leverage a biLSTM controller for configuration control to deal with the inaccurate configuration estimation from the internal sensing.
\item We carried out real experiments to validate the accuracy and versatility of our strategy. 
Offline tasks, including position control, orientation control, and obstacle avoidance, have been achieved successfully. Moreover, online interactions with the target and obstacle are performed.
\end{enumerate}
}

The remainder of the paper is structured as follows: 
Section \ref{sec2} describes the robot setup, the definitions of configuration and state, the configuration estimation approach, and data collected from our MSRA.
Section \ref{sec3} details the task space planning and control strategy S2C2A, covering configuration planning method S2C using biLSTM and configuration controller based on inaccurate internal sensing feedback. 
Section \ref{sec4} presents the real experimental results, which demonstrate that our controllers outperform our previous approach \cite{23QG} and can perform a wide range of tasks, from basic ones like position and orientation control to more complex ones like obstacle avoidance and online target following. 
Section \ref{sec5} summarizes this paper and discusses potential directions for future work.

\section{Setup and Specification}
\label{sec2}

\begin{figure*}[!ht]
\centering
\includegraphics[width=7.1in]{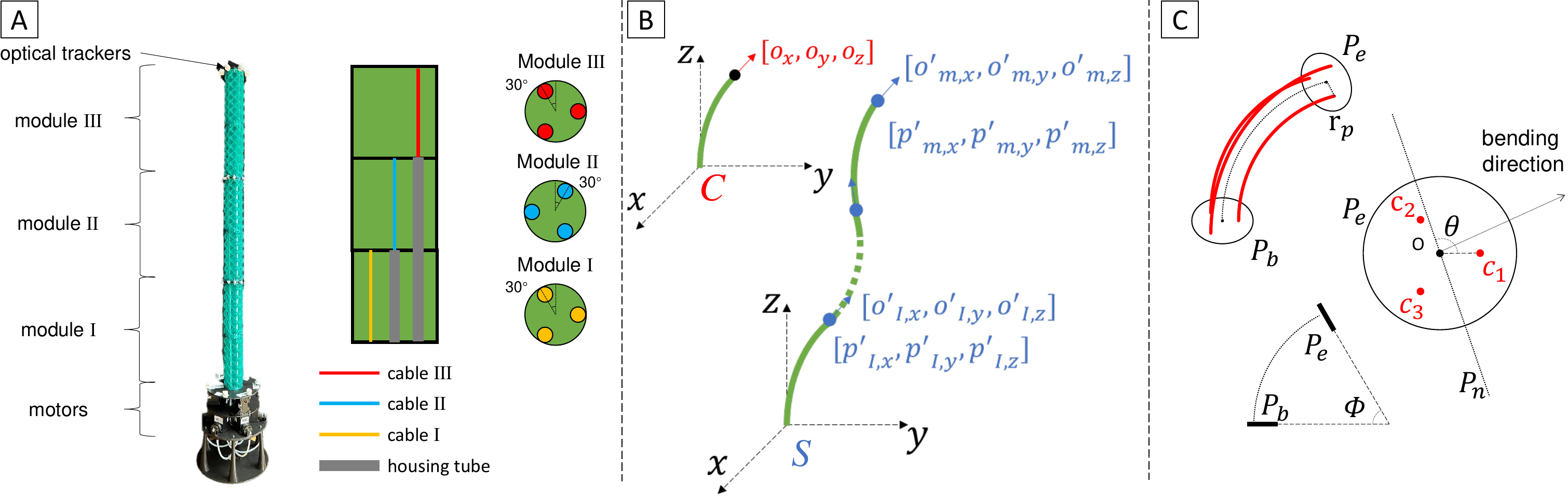}
\caption{
\red{(A) Robot setup. The cable-driven robot is composed of three independent segments, while each segment is actuated by three motors. The encoder in each motor provides feedback for configuration estimation. The optical tracking system is applied to provide ground truth robot states and configurations.
(B) Configuration and state. Configurations (red) refer to segment end orientation relative to the segment base, and states (blue) refer to the position and orientation of every segment relative to the robot base.
(C) Module actuation. The module configuration can be calculated by module bending angle $\phi$ and neutral surface direction $\theta$, which can be estimated by the displacements of cables $c_1, c_2, c_3$.}}
\label{fig2}
\end{figure*}

\subsection{Experimental Setup}
\label{sec2.1}
In this paper, we utilized a cable-driven modular soft robot arm for our experiments, as depicted in Fig. \ref{fig2}(A). \red{The robot comprises three independent modules (approximately 0.2 meters per module) based on Trimmed Helicoid (TH) structures. Each module is actuated by three motors (DYNAMIXEL xl330-m288-t) mounted on the MSRA base and connected via three cables arranged at 120° intervals along the circumferential direction. 
In this paper, we apply $I, II, III$ to represent the module number and $1\sim5$ to represent the time step.
The cables of Modules I and III are positioned at 0°, 120°, and 240°, while the cables of Module II are positioned at 60°, 180°, and 300°. The cables of Module I are directly attached to the motors with 20mm diameter pulley wheels. The cables of Modules II and III traverse housing tubes ($4\times2$mm PTFE tubes), thereby guaranteeing each module's independence. }

\red{
The housing tube-cable structure in our robot shares the same working principle as the bicycle brake system.
The cable inside can transmit the displacement between two ends while not deforming the outer tube.
Similarly, the housing tubes in our MSRA are fixed at both ends—extending from the base of the motor to the base of each module—while passing through pre-designed channels embedded within the TH structure. 
The force exerted by the internal tendons on the housing tubes is counterbalanced at both ends by the compressive force at the corresponding sections. 
As a result, the housing tubes effectively transmit the tension force directly from the motor base to a specific soft robotic section. 
Consequently, the tendon force influences only the targeted section without affecting other sections.
The only external factors influencing adjacent sections are the friction between the outer surface of the housing tube and the internal channels of the robot arm, as well as the reactive bending moment induced by arm flexion. 
However, these effects are negligible in our design due to the use of smooth and highly flexible PTFE tubes as housing tubes. 
Further details regarding the robot setup, including the MSRA material and mechanical properties, can be found in \cite{23QG} and the Supplementary Video.
}

For capturing a ground truth for the robot's position, several optical track markers are fixed at the end of each module and robot base.
Six optical tracking cameras (Optitrack Prime 13) and motion capture software Motive (Optitrack) collect each module end positions and orientations for NN training and only monitoring during control. A laptop (Ubuntu 20.04, CPU i5-12500H, and GPU RTX 3050) receives optical tracking information from the Motive computer via an ethernet switch and communicates with motors via one u2d2 power hub board (DYNAMIXEL). 
The whole system runs at about 10Hz in data collection and control tasks. 

\subsection{State and Configuration}
\label{sec2.2}
As shown in Fig. \ref{fig2}(B), we utilized the unit end orientation vector of each module subject to the module end $[o_x\ o_y\ o_z]\in R^{3}$ to represent module configuration $C$. 
We utilize it to describe module configuration following \cite{24ZCza} rather than bending directions and angles because such a representation is more suitable for NN data training as each element falls into the range of [-1, 1] and changes continuously during the MSRA motion. 
In contrast, bending direction angles undergo sudden shifts from 360$^\circ$ to 0$^\circ$ upon completing a full rotation, making them unsuitable for NN data training.

\red{The robot state $S=[S_I\ S_{II}\ \dots S_m]\in R^{6\times m}$ is composed of the states of all modules as shown in Fig. \ref{fig2}(B), where $m$ represents module number and is 3 in this work. For the module $n$, its state $S_n = [p'_n\ o'_n]^T \in R^{6}$ can be represent by its end position $p'_n=[p'_{n,x}\ p'_{n,y}\ p'_{n,z}]\in R^3$ and orientation $o'_n = [o'_{n,x}\ o'_{n,y}\ o'_{n,z}]\in R^3$ subject to the MSRA base. 
Overall, module configurations $C$ are independent, and from the geometry perspective, the configuration of one module does not affect the configuration of the other modules. 
In contrast, the state of one module is affected by the configurations of base modules and its own configuration. 
For instance, the state of the module $n$, $S_n$, is determined by the configurations of the modules $1\sim (n-1)$ and its own configuration.}

\subsection{Configuration Estimation}
\label{sec2.3}

To collect accurate MSRA states and configurations, most works use external sensing systems like the optical tracking system shown in Fig. \ref{fig2}(A) or electromagnetic tracking systems in \cite{16AA} for accurate feedback control. 
However, such systems have strict requirements on optical or electromagnetic environments and are unsuitable for real applications like agriculture and domestic environments. In this work, we aim to perform MSRA state control by only employing rough internal sensing feedback. Encoders in the motors are employed in this work, and some other internal sensors such as IMU\cite{24GP} and flex sensor\cite{17GG} may also be utilized for correcting the encoder and providing internal sensing feedback in future work.

The encoder in the motor can measure cable displacement and serve as an internal sensor. 
\red{We reconstruct the configuration based on the PCC model because it is one of the simplest soft robot models and produces a clear mapping from the cable displacement to the module configuration.}
Following \cite{20CS}, we estimate module orientation based on the cable displacement. Three cables $c_1, c_2, c_3$ are placed in each module, and $O$ represents the plane center as shown in Figure \ref{fig2}(C). 
In the initial state, the length of each cable and the module is $l_0$. During the motion, we define the cable lengths as $l_1, l_2, l_3$ and the cable displacement $a_i = l_i - l_0$, where $i=1, 2, 3$.
\red{Considering the low axial direction compressibility of each module, we assume that the module length does not change,  and the displacements follow:}
\begin{equation}
\label{eq3_0}
\begin{split}
a_1 + a_2 + a_3 = 0.
\end{split}
\end{equation}
To estimate the module configuration, first, we estimate the bending angle $\phi$ between the base plane $P_b$ and the end plane $P_e$ along with the angle $\theta$ between the neutral surface $P_n$ and the $O-c_1$ axis by
\begin{equation}
\label{eq3_1}
\begin{split}
\phi &= \frac{1}{r_p}\sqrt{a_1^2+\frac{(a_2-a_3)^2}{3}},\\
\theta &= atan2(-a_1, \frac{a_2-a_3}{\sqrt{3}}),\\
\end{split}
\end{equation}
where $r_p$ represents the distance between the plane center and the cable position, as shown in Fig. \ref{fig2}(C). Then, the configuration $[o_x,o_y,o_z]$ can be denoted as
\begin{equation}
\label{eq3_2}
\begin{split}
o_x &= sin(\phi) cos(\theta - \frac{\pi}{2}),\\ 
o_y &= sin(\phi) sin(\theta - \frac{\pi}{2}),\\ 
o_z &= cos(\phi).
\end{split}
\end{equation}
In this case, we can estimate the module configuration $[o_x,o_y,o_z]$ based on encoder feedback $a_1, a_2, a_3$.

\red{
Of note, the encoder cannot provide accurate configuration estimations due to cable slack, friction, and a possible mismatch between the real robots and the PCC model. 
In this work, we do not aim to address this error.
Instead, we bridge the mapping between this internal configuration space and the task and actuation space, as shown in Fig. \ref{fig1}(C).
Using this kind of rough and inaccurate configuration, our strategy can achieve various MSRA state control tasks thanks to our NN approaches.
In future work, we may only utilize raw feedback instead of estimation based on PCC.
}

\subsection{Data Collection}
\label{sec2.4}

\begin{figure*}[!ht]
\centering
\includegraphics[width=7.1in]{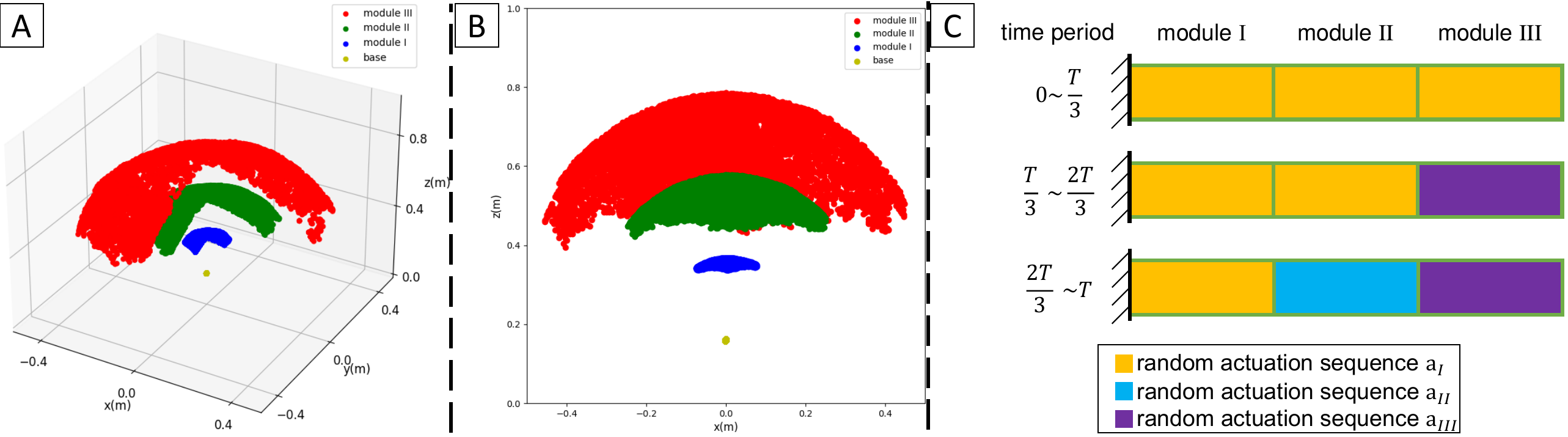}
\caption{
\red{(A) Some collected samples. Samples with $x>0$ and $y<0$ are hidden for visualization.
The positions of the base module end, middle module end, and end module end are depicted as blue, green, and red dots. The yellow dot represents the MSRA base.
(B) Collected samples exhibited on the x-z plane. 
}
(C) Data collection strategy specifically for MSRAs.
}
\label{fig3}
\end{figure*}

\red{When collecting data from MSRA, the purely motor babbling strategy applied in most neural network works \cite{23ZCb} may result in data samples confined to a limited space. 
To explore a larger area and reach the boundaries of the MSRA working space, all modules must bend in the same direction. 
However, this scenario is rare when module actuations are independent and random by the purely motor babbling strategy, especially with the increase of the module number.}
\red{In this case, we utilize a data collection strategy specifically for MSRA following \cite{24ZCza}, as shown in Fig. \ref{fig3}(B).}
All the modules are actuated with the same random action sequence $a_{I}$ in the first one-third period. Then, in the second one-third period, the base and middle modules are actuated with the same random action sequence $a_{I}$ while the end module is actuated by a different action sequence $a_{III}$. In the final one-third period, all these modules are actuated by different random action sequences $a_{I}, a_{II}, a_{III}$.
\red{Leveraging our data collection strategy, the MSRA can reach the working space edge in the beginning while exploring different configuration combinations after that.}

\red{Using the experiment devices mentioned in Sec. \ref{sec2.1}, we collected 9000 data points, as depicted in Fig. \ref{fig3}(A), and the working space length of the MSRA end on the x, y, and z axis are about 0.92 m, 0.92 m, and 0.39 m.
Meanwhile, the samples collected by the pure motor babbling strategy are constrained in a $0.87m\times 0.87m\times 0.34m$ space, which is smaller than ours.
This data collection approach specifically for MSRAs has demonstrated that it can explore a larger working space and variegated module configuration combinations both on our robot and in \cite{24ZCza}.}

\begin{figure}[!ht]
\centering
\includegraphics[width=3.4in]{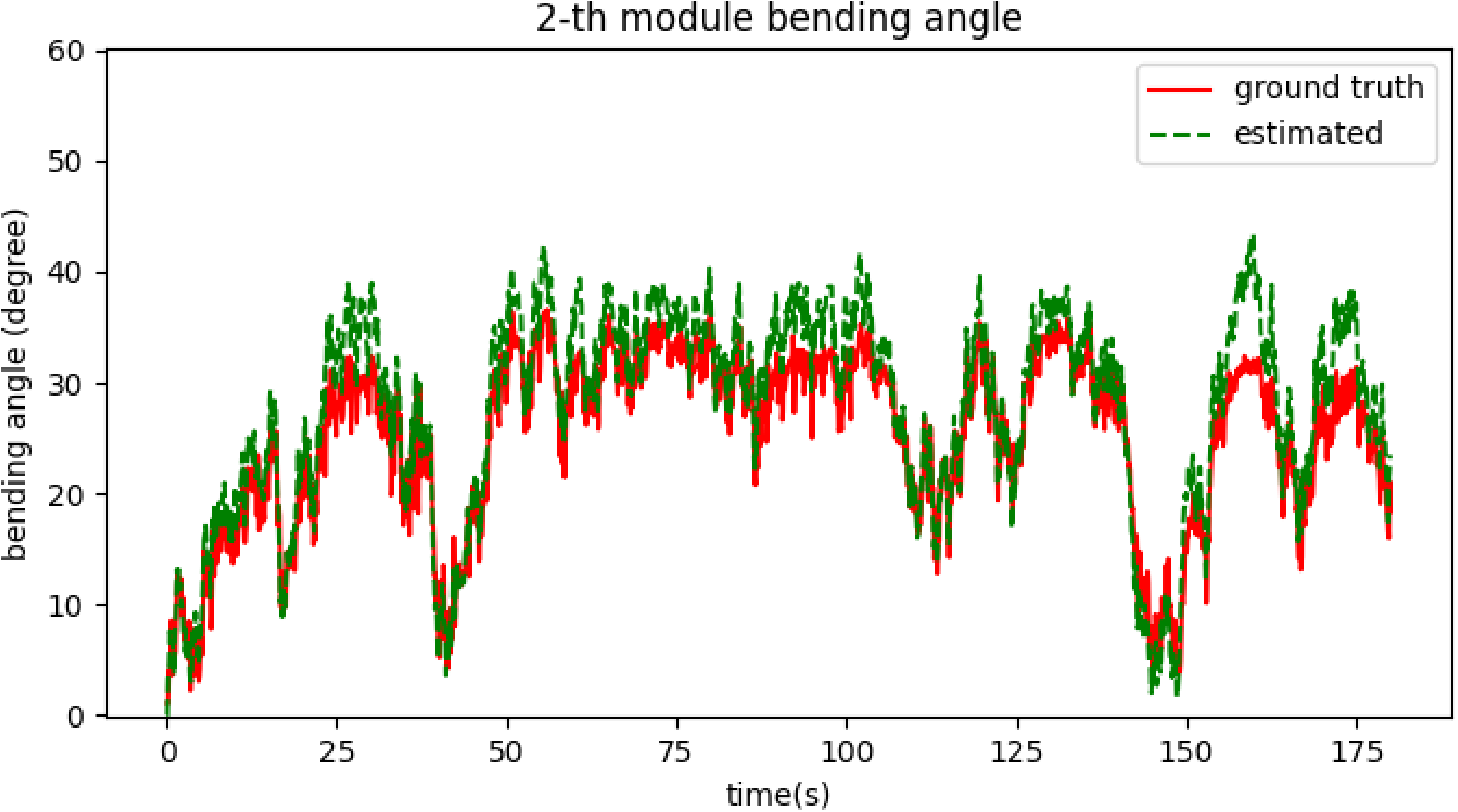}
\caption{
The middle module bending angles estimated by the optical tracking system (ground truth, red) and encoders in motors (green).
}
\label{fig4}
\end{figure}

\red{Based on the collected data, we compare the real module configurations collected by the optical tracking system (ground truth) and the configurations estimated by encoder feedback following Equation \ref{eq3_2}. The average estimation errors of the base, middle, and end modules are 2.6$^{\circ}$, 5.1$^{\circ}$, 8.9$^{\circ}$, and the maximal errors are 8.8$^{\circ}$, 19.6$^{\circ}$, 36.3$^{\circ}$, respectively.}
By taking the middle module as an example, the comparison of the ground truth and estimated module bending angles during dynamic motion is shown in Fig. \ref{fig4}.
The increase in error along the module sequence could be attributed to gravity, cable slack, and friction between the cables and modules. 

\red{
In this work, instead of using the optical tracking system as the configuration sensor or adjusting the estimation error, we focus on achieving state control even under the inaccurate configuration estimation, as shown in Fig. \ref{fig1}(C). 
Such an implementation endows the MSRA with a tracking ability that relies only on internal sensing feedback instead of costly external sensing systems like optical and electromagnetic tracking systems.
Internal configuration correction may be included in future work.
}

\begin{figure*}[!ht]
\centering
\includegraphics[width=7.1in]{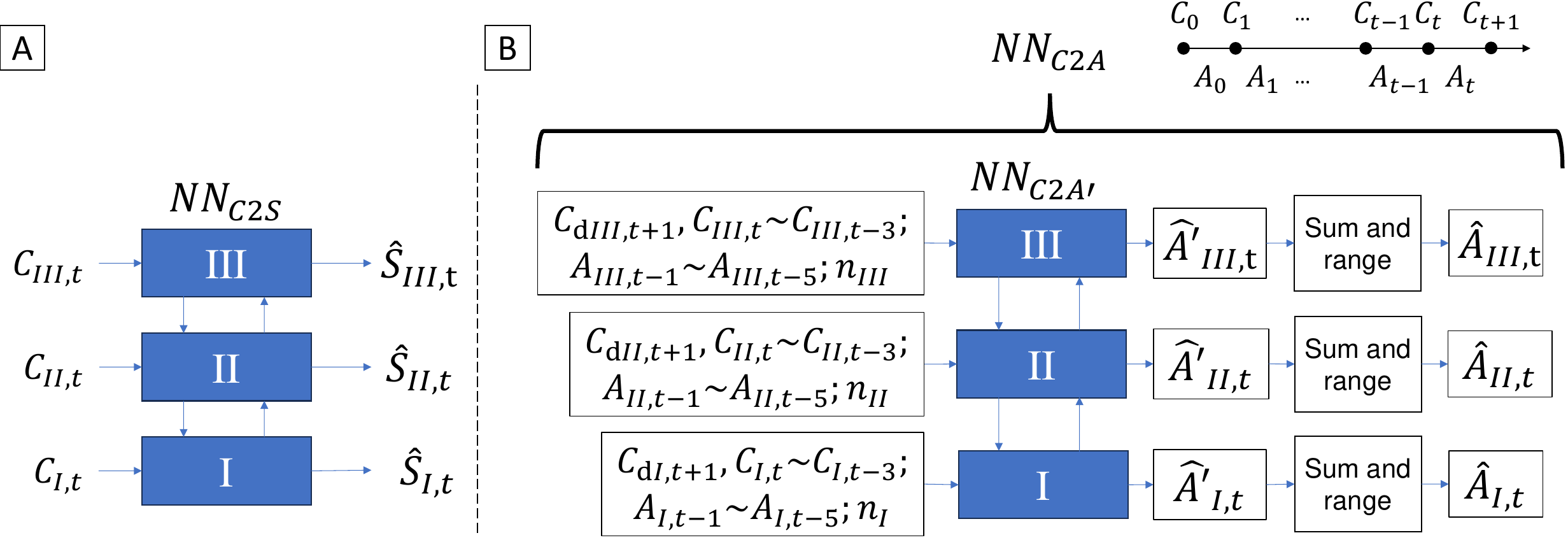}
\caption{
(A) Space sequence biLSTM network $NN_{C2S}$. Each biLSTM unit takes configurations $C_*$ of one module as input to estimate the robot states $\hat{S_*}$.
\red{(B) Configuration Controller $NN_{C2A}$. A biLSTM network $NN_{C2A'}$ decides the action $\hat{A_{*,t}}$ based on the previous actions $A_{*,t-1\sim t-5}$, target and previous states $C_{d*,t+1}, C_{*,t\sim t-3}$, and module labels $n_*$. The 'sum and range' layer is applied to guarantee the actuation sum and range.}
}
\label{fig5}
\end{figure*}

\section{Planning and Control Methods}
\label{sec3}

\subsection{Configuration Planning - S2C}
\label{sec3.1}
Our planning strategy is inspired by model predictive control (MPC), which determines action based on a forward model and an optimization problem. 
By adjusting the cost function in the optimization problem, diverse tasks can be performed. 
Considering the nonlinearity, hysteresis, and modularity of MSRAs, it is challenging to build an accurate physical model. 
Hence we leverage a biLSTM network $NN_{C2S}$ as the forward model estimating state $[\hat{p'}\ \hat{o'}]^T$ based on module configuration $C$, as shown in Fig. \ref{fig5}(A). 
All the data, including robot states, configurations, and actions, are rescaled to $[-1,1]$ for NN training.

In order to fulfill various tasks, a series of losses is integrated into the cost function.
To achieve position tracking and orientation tracking tasks, we introduce the losses 
\begin{equation}
\label{eq4_1}
\begin{split}
L_p &= {\Vert p'_d-\hat{p'} \Vert}_2,\\
L_o &= {\Vert o'_d-\hat{o'} \Vert}_2,
\end{split}
\end{equation}
where $p'_d, o'_d$ are the desired position and orientation.

To achieve obstacle avoidance, we define the loss
\begin{equation}
\label{eq4_2}
\begin{split}
L_{ob} &= 
\begin{cases}
\frac{1}{d}, & d \leq r,\\
0,   & d > r,\\
\end{cases}\\
\end{split}
\end{equation}
where $d = {\Vert p'_{ob}-\hat{p'} \Vert}_2$ refers to the distance between the estimated position $\hat{p'}$ and the obstacle center $p'_{ob}$. \red{$r$ defines the obstacle avoidance threshold according to the obstacle risk level and obstacle size. In future work, we may update the obstacle center and avoidance threshold online for adjusting to the changing and moving obstacles.}

Finally, we introduce a configuration change loss $L_d$ to constrain the configuration change between two continuous steps.
\begin{equation}
\label{eq4_4}
\begin{split}
L_d &= {\Vert \triangle C \Vert}_2,\\
\end{split}
\end{equation}
where $\triangle C = C - C_0$ denotes the difference between the last configuration $C_0$ and the next configuration $C$.

Overall, the configuration planning strategy can be illustrated as 
\begin{equation}
\label{eq4_5}
\begin{split}
{\min_{C}}\ \mu_p L_p +&\mu_o L_o + \mu_{ob} L_{ob} + \mu_{d} L_{d},\\
{s.t.}\ [\hat{p'}\ \hat{o'}]^T&=NN_{C2S}(C)\\
\end{split}
\end{equation}
where $\mu_p, \mu_o, \mu_{ob}, \mu_d$ represent the weights of position control, orientation control, obstacle avoidance, and configuration change. $NN_{C2S}$ serves as a forward model estimating the robot state $[\hat{p'}\ \hat{o'}]^T$ from the module configurations $C$, and the module configurations $C$ are optimized to minimize the cost function. 
\red{The loss weights can be adjusted according to task requirements, like the task propriety order; those applied in the following experiments have been decided after a trial and error process.}

Compared to other optimization works using physical models mentioned in Sec. \ref{sec1}, our data-driven strategy can reflect the real motion features of the MSRA. 
Compared to other NN implementations in soft robots, which only target a single task, such a planning strategy is adaptive to various tasks. 
Moreover, this planning strategy can be implemented both offline and online, depending on the task requirements.

\red{
Before the planning, $NN_{C2S}$ is trained using the dataset in Sec. \ref{sec2.4}.
To minimize the cost functions in Equation \ref{eq4_5} and generate the target configuration online or offline, the $NN_{C2S}$ serves as a forward model, and its parameters are frozen during optimization.
We employ Pytorch\cite{19AP} for planning, and the optimizer is Adam. The learning rate is 0.02, and the optimization iteration is 10.}

\subsection{Configuration Control - C2A}
\label{sec3.2}

Following \cite{24ZCza}, we utilize a biLSTM $NN_{C2A'}$ and a 'sum and range' layer to compose $NN_{C2A}$ as the configuration controller for MSRAs, as shown in Fig. \ref{fig5}(B). 
The controller can be represented as
\begin{equation}
\label{eq4_6}
\begin{split}
\hat{A_t} = NN_{C2A}(C_{d,t+1}; C_t\sim C_{t-3}; A_{t-1}\sim A_{t-5};n),
\end{split}
\end{equation}
where $NN_{C2A}$ illustrates the controller that determines the $t$ step action $\hat{A_t}$ based on the target configuration $C_{d,t+1}$, previous configurations $C_t\sim C_{t-3}$, previous actions $A_{t-1}\sim A_{t-5}$, and module label $n$. 

The module label $n_m\in[-1,1]$ of the $m$-th module is utilized to infer the module position in the module sequence, which can be presented as
\begin{equation}
\label{eq4_7}
\begin{split}
n_m = \frac{2(m-1)}{n_{sum}-1}-1,
\end{split}
\end{equation}
where $n_{sum}$ denotes the amount of modules in the MSRA. A higher label represents that the module is near the MSRA end rather than the base. 

Considering the low axial compressibility of our module, the actions of each module $\hat{A_{m,t}}\in R^3$ should follow the constraint $\sum_{i=1}^3 \hat{A_{i,t}} = 0$ to maintain the central axis length invariant following Equation \ref{eq3_0}. Also, they should be in the range $[-1,1]$ for NN training. Hence, one layer follows the biLSTM 
\begin{equation}
\label{eq4_8}
\begin{split}
\hat{A_{m,t}} = \frac{3 \tanh(\hat{A_{m,t}'}) - \sum_{i=1}^3 \tanh(\hat{A_{i,t}'})}{4},
\end{split}
\end{equation}
where $\hat{A_{m,t}'}$ represents the biLSTM $NN_{C2A'}$ output of the $m$-th module at $t$-th step.

\red{Overall, as shown in Fig. \ref{fig5}(B), the target configuration $C_{dm,t+1}$, previous configurations $C_{m,t}\sim C_{m,t-3}$, previous actions $A_{m,t-1}\sim A_{m,t-5}$, and module label $n_m$ are formatted into a 1-D vector and fed into the biLSTM unit $m$ as input, where $m=1,2,3$ represent the module number. 
The output of the unit $m$ is the estimated actuation $\hat{A_{m,t}}$ to reach the target configuration $C_{dm,t+1}$ considering the previous module configurations, actuations, and hidden states from the other units.
BiLSTM is chosen as the configuration controller because it outperforms some other neural network controllers. The details of the biLSTM controller and the comparison with the other neural network controllers can be found in our previous work \cite{24ZCza}.}



Utilizing the dataset collected in Section \ref{sec2.4}, we train $NN_{C2S}$ and $NN_{C2A}$ for configuration planning and control, and the NN hyperparameters are shown in Table \ref{table2}. Of note, the configuration applied in NN training is only from internal sensors, not the optical system.
\red{
The early stopping strategy is applied to avoid overfitting.
These hyperparameters are chosen after trial and error, and  $NN_{C2A}$ contains fewer layers and a smaller hidden state size to decrease the online computation burden.}

\begin{table}[!ht]
\caption{NN hyperparameters}
\centering
\begin{tabular}{l|l l}
&$NN_{C2S}$&$NN_{C2A}$\\
\hline
layer number&4&2\\
time step&/&5\\
hidden state size&64&32\\
learning rate&0.001&0.001\\
batch size&64&64\\
optimizer&\red{Adam}&Adam\\
average estimation error &1.93$\pm$1.87\%&0.91$\pm$1.90\%\\
\end{tabular}
\label{table2}
\end{table}

\section{Experimental Results}
\label{sec4}

In this section, we demonstrate the versatility of our planning and control strategy in various tasks. 
\rred{First, we validate some of our strategies with primary experiments in simulation, including the data collection strategy and the robustness to sensing and actuation coupling in Sec. \ref{sec4.0}. Then,} we compare our approach with our previous planning and controller strategy applied to our robot in \cite{23QG}. The comparison results in position and orientation control are shown in Sec. \ref{sec4.1} and \ref{sec4.2}. Then, we validate our approach in some high-level tasks, like position constraint in Sec. \ref{sec4.3}, obstacle avoidance in Sec. \ref{sec4.4}, and online interaction in Sec. \ref{sec4.5}.

\subsection{\rred{Simulation Validation}}
\label{sec4.0}

\begin{figure*}[!ht]
\centering
\includegraphics[width=7.1in]{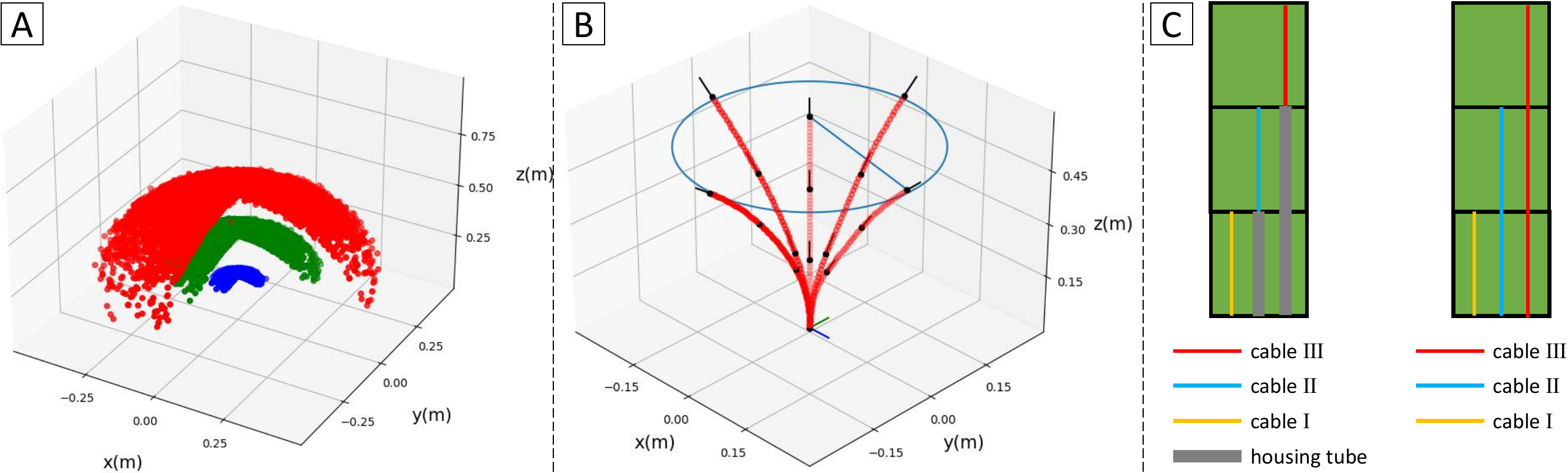}
\caption{
\rred{
(A) The dataset collected by our strategy in the simulation. The green, orange, and blue dots represent the end positions of the base, middle, and end modules.
(B) The target MSRA end trajectory (blue) and the MSRA motion. Each module end is represented by one black dot, and its orientation is represented by a black vector.
(C) The independent (left) and coupled (right) actuation diagram.
}
}
\label{figsim1}
\end{figure*}

\rred{
We apply a soft robot simulator based on the pseudo-rigid model as reported in \cite{18CS}, as shown in Fig. \ref{figsim1}(B). 
A three-module MSRA is built in this simulator, and the length of each module is 0.2m. To validate the generalization of our strategies, four cables, rather than three cables in the real robots, are simulated in the robot, and the actuation value is force, not cable displacement.
}

\subsubsection{\rred{Data Collection Strategy Comparison}}
\begin{figure}[!ht]
\centering
\includegraphics[width=\linewidth]{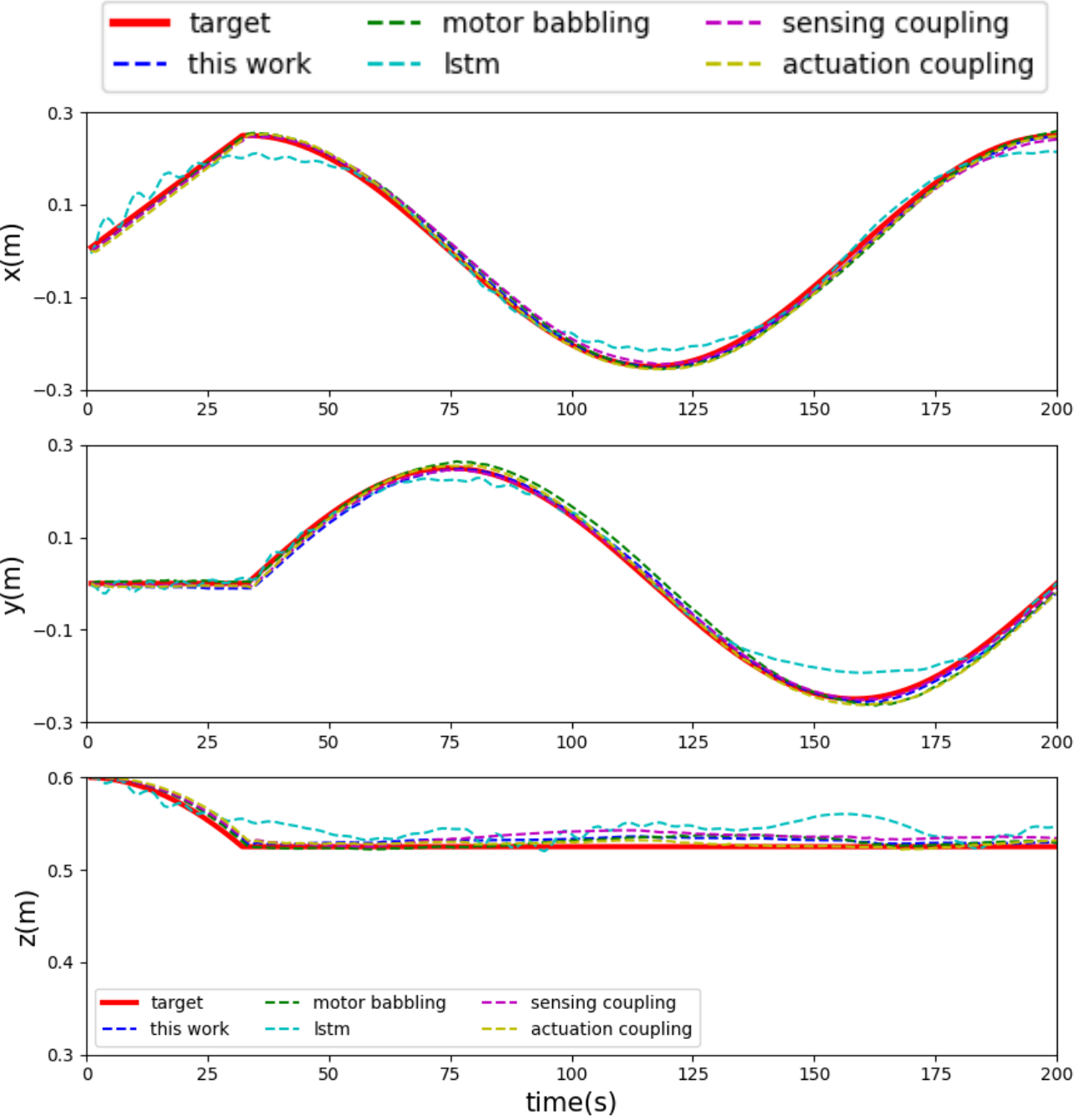}
\caption{   
\rred{
The target (red) and measured motion of MSRA ends on the x, y, and z axes applying our control strategy (blue), motor babbling strategy (green), LSTM (light blue), under sensing coupling (purple) and actuation coupling (yellow).
}
}
\label{figsim2}
\end{figure}

\rred{
In Sec. \ref{sec2.4} and our previous work \cite{24ZCza}, we have proven that our data collection strategy produces a larger working space than the common motor babbling strategy in MSRA. In this subsection, we also test the influence on control accuracy. 
First, we employ our strategy, including data collection strategy in Fig. \ref{fig1}(C), S2C planning, and C2A control, in simulation. 
9000 samples are collected, as shown in Fig. \ref{figsim1}(A). 
The error of this work is shown in Table \ref{tablesim1}. The MSRA motion is shown in Fig. \ref{figsim1}(B) and Fig. \ref{figsim2}. The low error demonstrates that our strategy can achieve position control in MSRA.}

\rred{Then, we apply the dataset collected by the common motor babbling strategy for neural network training, including $NN_{C2S}$ and $NN_{C2A}$. These two networks are applied for position control and named 'Motor babbling' in Table \ref{tablesim1}.
The lower error of our strategy demonstrates that the dataset collected by our strategy produces a larger working space and contributes to accurate control.
A dataset covering a larger working space provides richer configuration combinations, such as those reaching the edge, and enhances the neural network’s generalization ability to represent MSRA motion.
}

\begin{table}[!ht]
\caption{\rred{Average errors and standard deviations comparison in position control in simulation}}
\centering
\begin{tabular}{l|l}
Approach&Errors (cm)\\
\hline
This work          &$\bf{1.5\pm0.3}$\\
Motor babbling     &1.9$\pm$0.7\\
LSTM               &3.3$\pm$1.7\\
Sensing coupling   &1.7$\pm$0.3\\
Actuation coupling &1.6$\pm$0.5\\
\end{tabular}
\label{tablesim1}
\end{table}

\subsubsection{\rred{Data-driven Approach Comparison}}
\rred{We compare our strategy with another data-driven approach widely applied in soft robot control, which is LSTM \cite{23ZCb,22TT}. In the existing works, the previous robot states and control target (end position in this task) are the LSTM input, while the LSTM decides the next actuation. 
The error of LSTM is higher than that of this work, as shown in Table \ref{tablesim1}.}

\rred{
The NN applications in soft robotics are motivated by the specific challenges of the hardware. 
Multilayer perceptron is introduced in soft robotics to address the nonlinearity.
Then, RNN targeted at sequence problems is widely applied due to the delayed motion. 
Following the same principle, when meeting challenges about modularity in MSRA, we utilize a novel NN, biLSTM, in planning and control.
Unlike existing data-driven controllers, which apply target position as input to decide actuation, we utilize biLSTM as the forward model in optimization planning to cope with the complex state-to-configuration mapping caused by modularity.
}

\subsubsection{\rred{Sensing Coupling Robustness}}

\rred{Module coupling is one of the most significant challenges in MSRA control. 
Some sensors in MSRA, like the encoders, pass through the whole MSRA and are affected by the module coupling. In this case, we built a sensing coupling model, denoted as}
\begin{equation}
\label{eq4_sim1}
\begin{split}
\begin{bmatrix}
\hat{C_I}\\
\hat{C_{II}}\\
\hat{C_{III}}\\
\end{bmatrix}
=
\begin{bmatrix}
0.85 & 0.10 & 0.05\\
0.20 & 0.70 & 0.10\\
0.15 & 0.30 & 0.55\\
\end{bmatrix}
\begin{bmatrix}
{C_I}\\
{C_{II}}\\
{C_{III}}\\
\end{bmatrix},
\end{split}
\end{equation}
\rred{where $\hat{C_*}$ represent the inaccurate configuration suffered from sensing coupling and ${C_*}$ represent the accurate configuration.
We train our strategy with the inaccurate configuration and achieve position control, resulting in a similar error to 'Ours' as shown in Table \ref{tablesim1}. Similar errors demonstrate the robustness of our strategy against sensing coupling.
}
\subsubsection{\rred{Actuation Coupling Robustness}}

\rred{In addition to sensing coupling, some actuation units, like cables, also pass through the whole MSRA, like the old version of our real MSRA mentioned in \cite{23QG}. To test the robustness against actuation coupling, we couple the cables in simulation, as shown in Fig. \ref{figsim1}(C). Then, we collect a dataset with this kind of MSRA and train our strategy with the new dataset. The error is shown in Table \ref{tablesim1}, demonstrating that our strategy shows robustness against actuation coupling.}

\subsection{Position Control}
\label{sec4.1}
\begin{figure}[ht]
\centering
\includegraphics[width=3.4in]{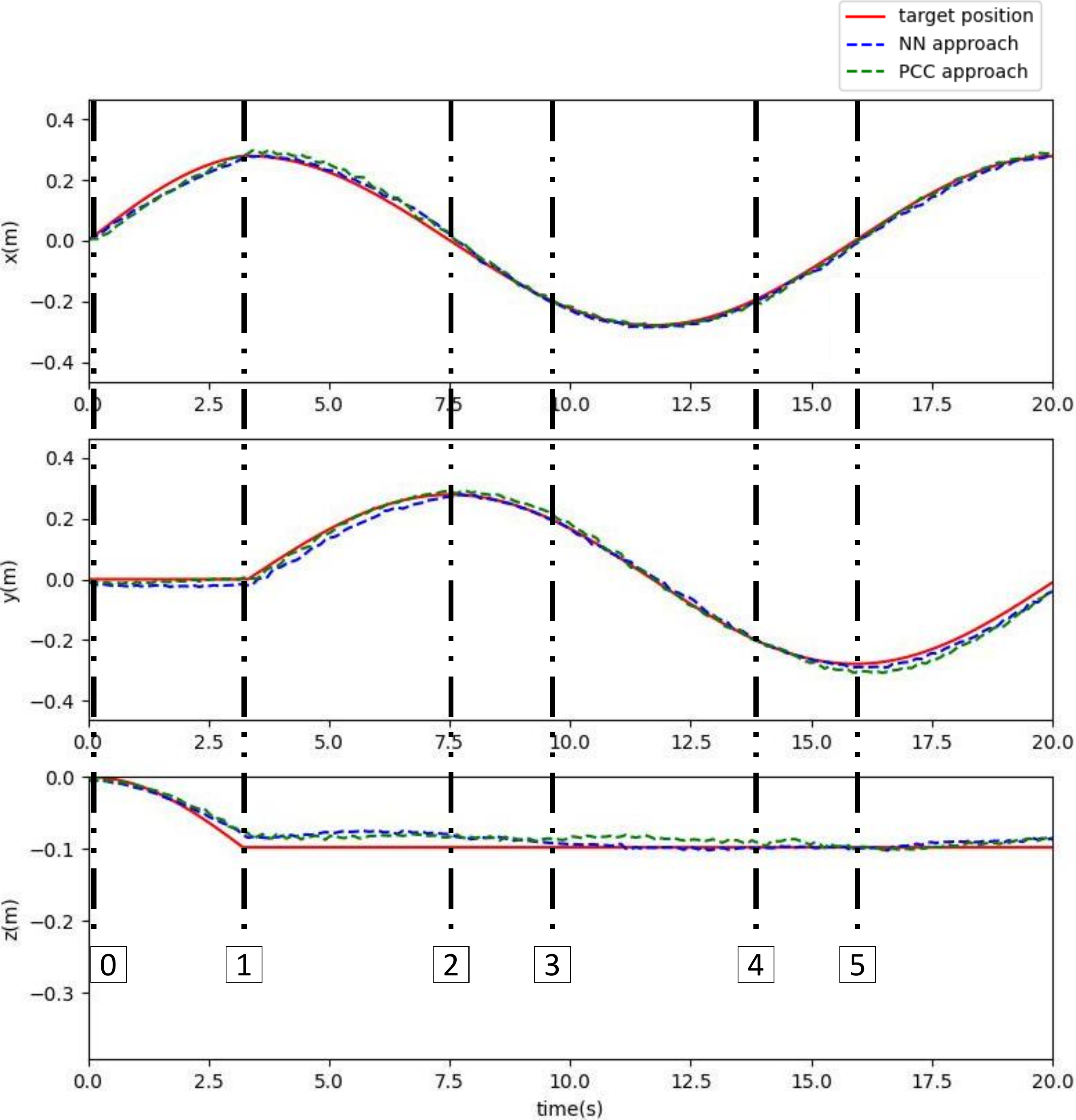}
\caption{The target (red) trajectories in the position control task and real trajectories applying our strategy (blue) and PCC strategy (green).}
\label{fig6}
\end{figure}

\begin{figure}[ht]
\centering
\includegraphics[width=3.4in]{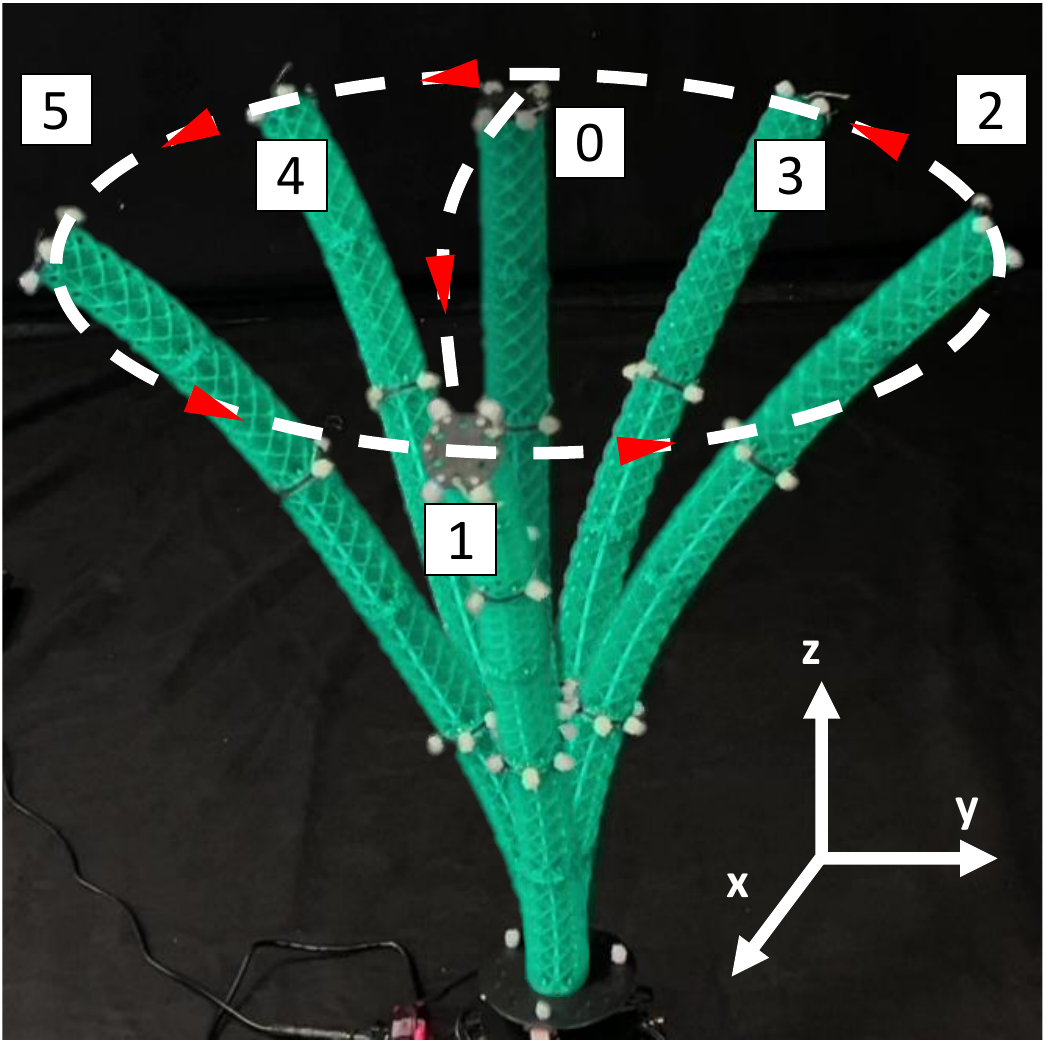}
\caption{\red{The MSRA motion in position control using the configuration controller $NN_{C2A}$. The MSRA first decreases the height (0-1) and draws a circle (1-2-3-4-5). The white dotted lines represent the MSRA end trajectories, and the red arrows represent the motion directions.}}
\label{fig7}
\end{figure}

We propose a position trajectory in the task space, as shown in the red lines in Fig. \ref{fig6} and the white dotted trajectory in Fig. \ref{fig7}, and employ our biLSTM approach and PCC approach in \cite{23QG} to achieve position control. Following the target trajectory, the MSRA will first bend along the x-axis and rotate to move along a circle.
We include the losses $L_p$ and $L_d$ for position tracking and compare the performance of the PCC strategy \cite{23QG} and our biLSTM strategy.
In this task, $u_p=1$ and $u_d=0.5$.
The errors of trajectories using the PCC and our strategies are shown in Table \ref{table3}.
The MSRA trajectories and robot motions are shown in Fig. \ref{fig6} and \ref{fig7}. All the experiment videos can be found in the Supplementary Video.

\begin{table}[!ht]
\caption{\red{Average errors and standard deviations comparison in position and orientation control}}
\centering
\begin{tabular}{p{58pt}|p{45pt} p{45pt} p{45pt}}
task&position error&orientation error (z)&orientation error (x)\\
\hline
p-PCC\cite{23QG}&2.5$\pm$0.8\ cm&/&/\\
p-ours&$\bm{2.2\pm1.1}$cm&/&/\\
\hline
40$^{\circ}$-PCC\cite{23QG}&3.3$\pm$1.0\ cm&2.1$\pm$1.4$^{\circ}$&/\\
40$^{\circ}$-ours&$\bm{2.9\pm0.8}$cm&$\bm{2.1\pm1.3}^{\circ}$&/\\
\hline
50$^{\circ}$-PCC\cite{23QG}&3.9$\pm$1.4\ cm&4.5$\pm$2.2$^{\circ}$&/\\
50$^{\circ}$-ours&$\bm{2.3\pm0.7}$cm&$\bm{2.8\pm1.4}^{\circ}$&/\\
\hline
60$^{\circ}$-PCC\cite{23QG}&4.5$\pm$1.0\ cm&7.5$\pm$3.7$^{\circ}$&/\\
60$^{\circ}$-ours&$\bm{2.2\pm1.1}$cm&$\bm{3.5\pm2.9}^{\circ}$&/\\
\hline
0$^{\circ} $-PCC\cite{23QG}&2.9$\pm$0.8\ cm&4.7$\pm$2.2$^{\circ}$&/\\
0$^{\circ} $-ours&$\bm{2.6\pm1.0}$cm&$\bm{3.3\pm1.0}^{\circ}$&/\\
\hline
50$^{\circ}$a20$^{\circ}$-PCC\cite{23QG}&4.8$\pm$1.4\ cm&3.9$\pm$2.3$^{\circ}$&6.8$\pm$3.4$^{\circ}$\\
50$^{\circ}$a20$^{\circ}$-ours&$\bm{3.1\pm1.6}$cm&$\bm{2.4\pm1.5}^{\circ}$&$\bm{3.6\pm2.7}^{\circ}$\\
\hline
50$^{\circ}$c20$^{\circ}$-PCC\cite{23QG}&3.8$\pm$1.1\ cm&5.9$\pm$2.0$^{\circ}$&4.4$\pm$2.2$^{\circ}$\\
50$^{\circ}$c20$^{\circ}$-ours&$\bm{3.1\pm1.2}$cm&$\bm{3.0\pm2.3}^{\circ}$&$\bm{3.5\pm3.1}^{\circ}$\\
\hline
\hline
Average-PCC\cite{23QG}&3.7$\pm$1.1\ cm&4.1$\pm$2.3$^{\circ}$&5.6$\pm$2.8$^{\circ}$\\
Average-ours&$\bm{2.6\pm1.1}$cm&$\bm{2.9\pm1.7}^{\circ}$&$\bm{3.6\pm2.9}^{\circ}$\\
\end{tabular}
\label{table3}
\end{table}

Of note, the MSRA states collected by the optical tracking system are only used for comparing with the target trajectory, and we do not utilize them as feedback in the following experiments, which is proven obviously in Section \ref{sec4.5}.
Only encoder feedback is fed into the configuration controller.
\red{The experimental results show that both the PCC approach and our approach can achieve relatively low errors ($\leq2.5cm$) in position control, and our strategy outperforms the PCC one \cite{23QG}.}

\subsection{Position and Orientation Control}
\label{sec4.2}
\begin{figure}[!ht]
\centering
\includegraphics[width=3.2in]{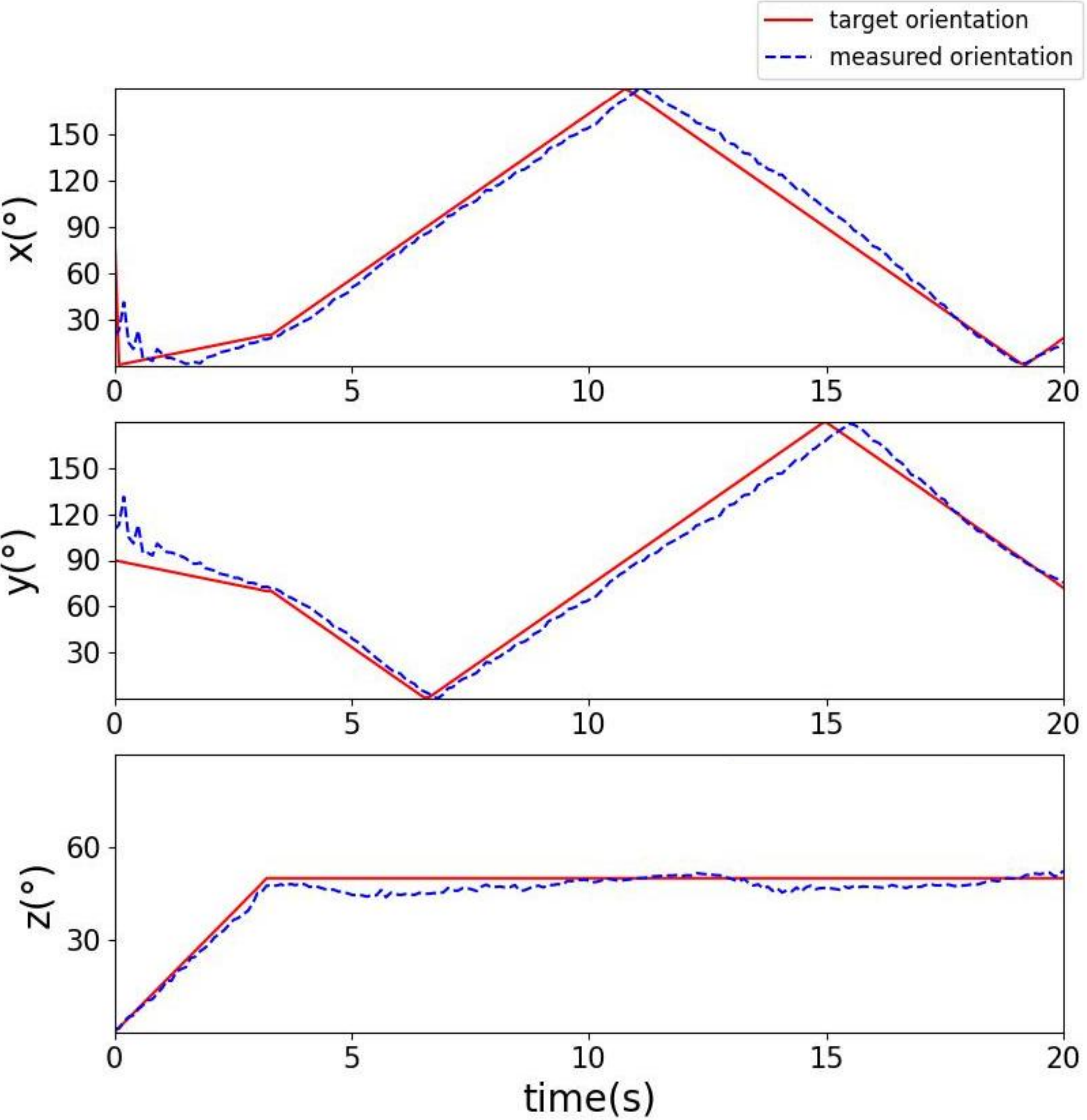}
\caption{The target (red) and measured (blue) orientation trajectories in the position and orientation control task for the task 50$^{\circ}$ anticlockwise 20$^{\circ}$.}
\label{fig8}
\end{figure}

In addition to the position trajectory shown in Fig. \ref{fig6}, we add target angle trajectories for MSRA end orientation, as depicted in the red lines in Fig. \ref{fig8}. In these tasks, we include the losses $L_p, L_d$ and the orientation loss $L_o$ for tracking. 

\begin{figure}[!ht]
\centering
\includegraphics[width=3.2in]{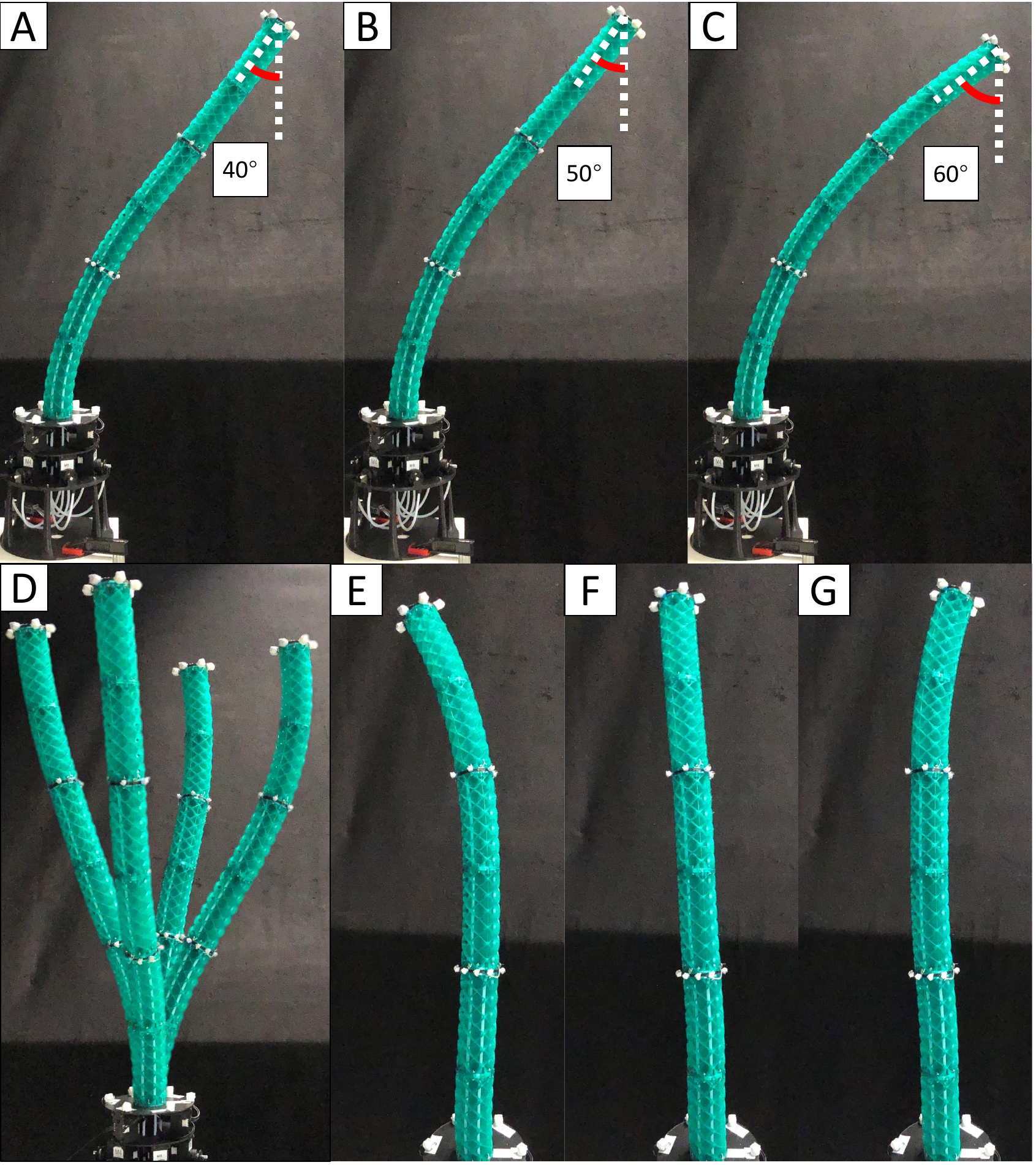}
\caption{The MSRA motion for (A) 40$^{\circ}$, (B) 50$^{\circ}$, (C) 60$^{\circ}$, (D) 0$^{\circ}$ in position and orientation control. MSRA motion comparison for the task (E) 50$^{\circ}$a20$^{\circ}$, (F) 50$^{\circ}$, and (G) 50$^{\circ}$c20$^{\circ}$.}
\label{fig9}
\end{figure}

First, we define three tasks requiring the bending angle to 40$^{\circ}$, 50$^{\circ}$, and 60$^{\circ}$, as shown in Fig. \ref{fig9}(A), (B), and (C). In the fourth task, the MSRA is controlled to keep the end upward while following a circle, as illustrated in Fig. \ref{fig9}(D). In these tasks, $u_p=1$, $u_d=0.5$, and $u_o=2$. The errors are shown in Table \ref{table3}. All the experiment videos can be found in the Supplementary Video.

\begin{figure*}[!ht]
\centering
\includegraphics[width=7.1in]{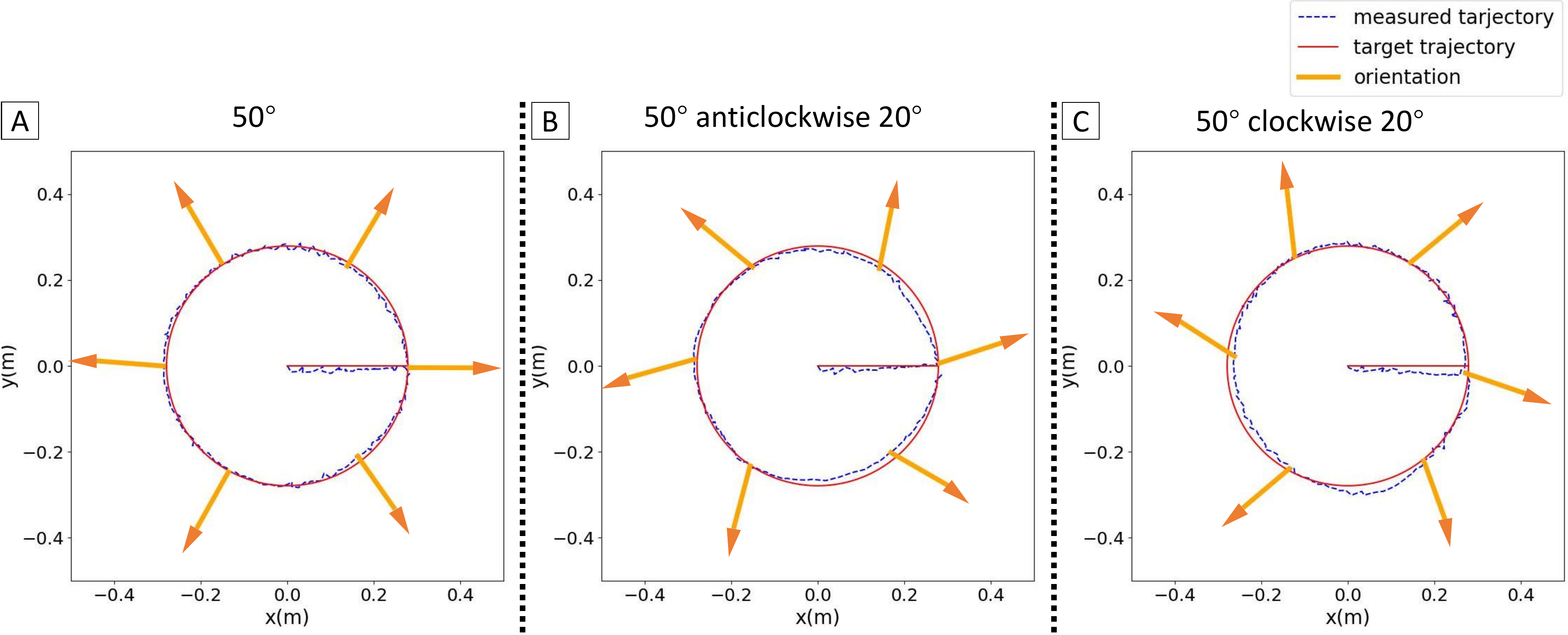}
\caption{The target (red) and real (blue) MSRA end motion on the x-y plane for the task (A) 50$^{\circ}$, (B) 50$^{\circ}$a20$^{\circ}$, (C) 50$^{\circ}$c20$^{\circ}$. The yellow lines depict the end orientations on the x-y plane.}
\label{fig10}
\end{figure*}

Based on the task 50$^{\circ}$, we also control the MSRA to bend 20$^{\circ}$ anticlockwise (50$^{\circ}$a20$^{\circ}$) and clockwise (50$^{\circ}$c20$^{\circ}$) on the x-y plane. The errors are shown in Table \ref{table3}. We show the MSRA deformation on the same end position in Fig. \ref{fig9}(E), (F), and (G). The MSRA orientations on the x-y plane in these tasks are shown in Fig. \ref{fig10}.
It is obvious that our approach can perform position and orientation control simultaneously and achieve satisfying accuracy even in some intricate tasks.
\red{The experimental results of the position and orientation control show that our strategy outperforms the PCC controller; hence, we employ it in the following high-level tasks.}

\subsection{Position Constraint}
\label{sec4.3}
\begin{figure}[!ht]
\centering
\includegraphics[width=3.4in]{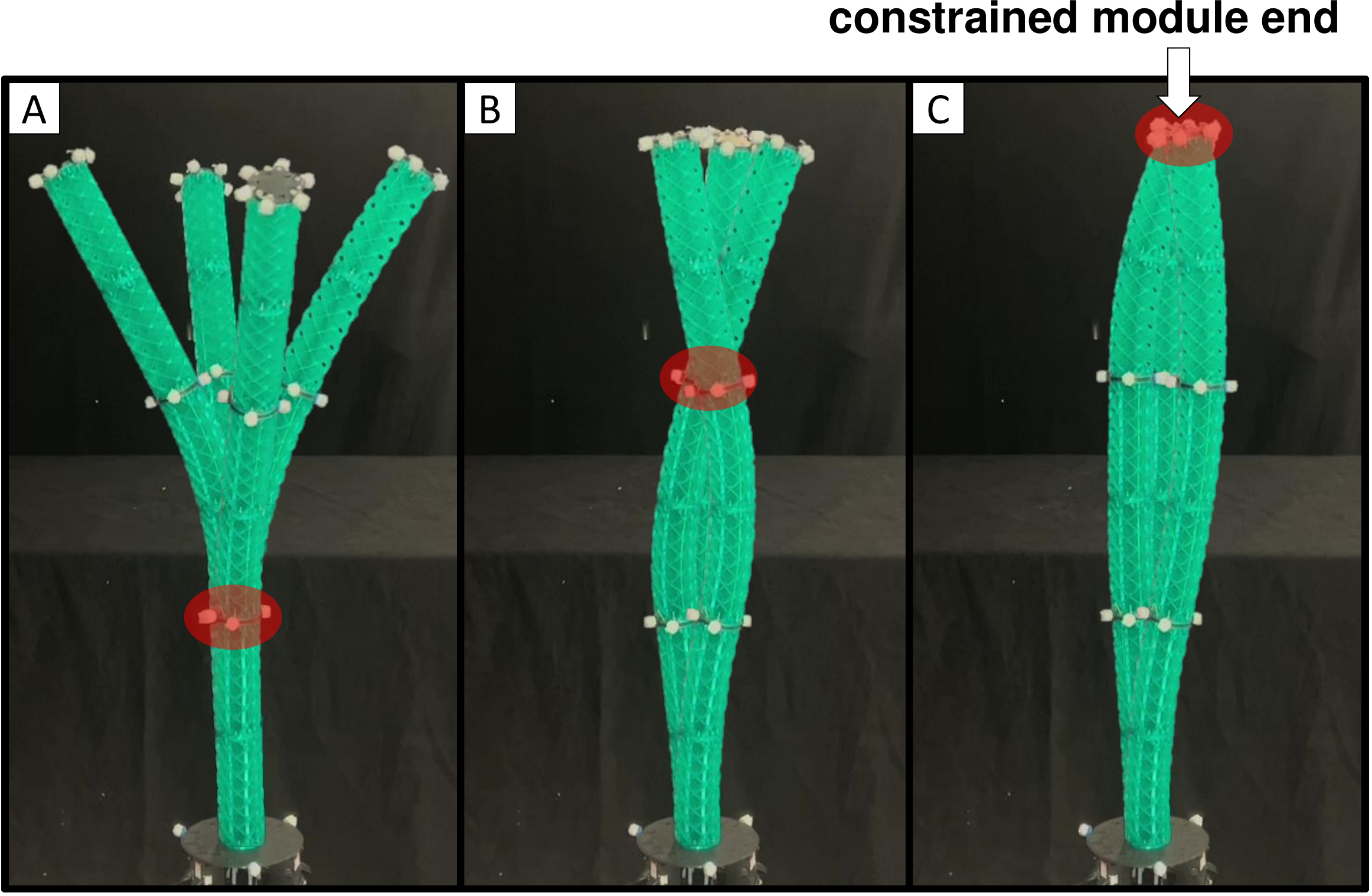}
\caption{The MSRA motion for the task (A) base, (B) middle, and (C) end in position constraint control. The red transparent circles depict the module end that remains its position.}
\label{fig11}
\end{figure}

In these tasks, we aim to keep the end position of the base, middle, and end module invariant during motion, as depicted in Fig. \ref{fig11}. 
We include the losses $L_p, L_d, L_o$ for this task. 
\red{In task 'base,' we control the MSRA end to follow a circle trajectory while maintaining the base module position invariant, as shown in Fig. \ref{fig11}(A). 
In tasks 'middle' and 'end,' we aim to keep the middle and end module end positions invariant and change the same module end orientations from 0$^\circ$ to 360$^\circ$, as shown in Fig. \ref{fig11}(B) and (C). }
The errors are shown in Table \ref{table5}. During these motions, the corresponding module end positions remain static, but the orientations keep changing for MSRA motion.
All the experiment videos can be found in the Supplementary Video.

\begin{table}[ht]
\caption{Average errors and standard deviations in position constraint}
\centering
\begin{tabular}{l|l l}
task&constraint error&motion error\\
\hline
base&0.2$\pm$0.1\ cm&3.6$\pm$1.7\ cm\\
middle&1.0$\pm$0.3\ cm&3.2$\pm$1.9$^{\circ}$\\
end&3.9$\pm$0.9\ cm&4.3$\pm$2.5$^{\circ}$\\
\end{tabular}
\label{table5}
\end{table}

\subsection{Obstacle Avoidance}
\label{sec4.4}

\begin{figure*}[!ht]
\centering
\includegraphics[width=7.1in]{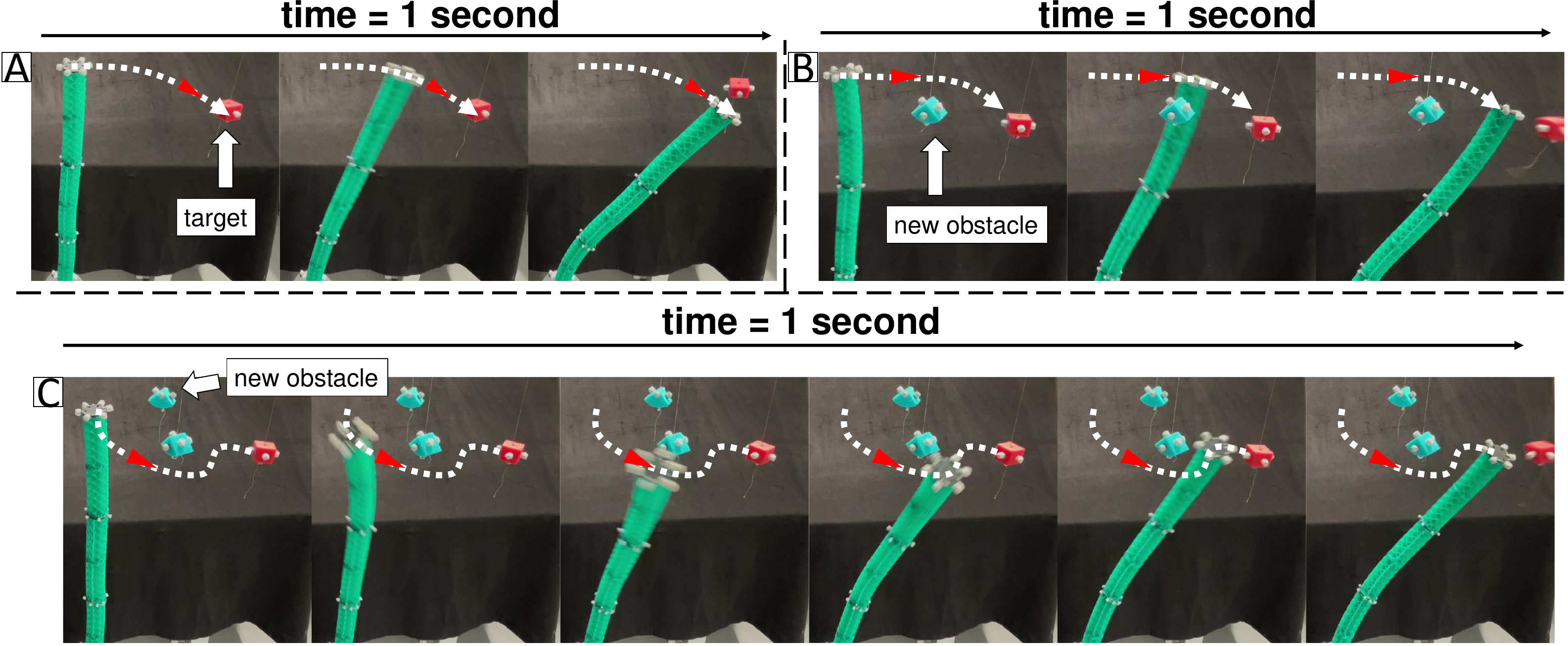}
\caption{The MSRA motion for the task in obstacle avoidance control with (A) 0, (B) 1, and (C) 2 obstacles. The white dotted lines represent the MSRA end trajectories, and the red arrows represent the motion directions. Red and blue sponge cubes represent the target and obstacles.}
\label{fig12}
\end{figure*}

\red{In obstacle avoidance, we include the losses $L_p, L_d$ for position control and leverage the obstacle loss $L_{ob}$ ($u_{ob} = 1$ in the following tasks).} First, we set a red sponge bar as the target and control the MSRA end to reach it, as illustrated in Fig. \ref{fig12}(A). Then, we put one blue sponge bar as the obstacle in the last MSRA end trajectory and include $L_{ob}$ for replanning. The MSRA end first moves away from the obstacle to avoid collision (Fig. \ref{fig12}(B2)) and then reaches the target (Fig. \ref{fig12}(B3)). Furthermore, we set one more obstacle in the last MSRA end trajectory as depicted in Fig. \ref{fig12}(C). In this case, the S2C planning approach changes the trajectory and helps the MSRA reach the target.

\begin{figure*}[!ht]
\centering
\includegraphics[width=7.1in]{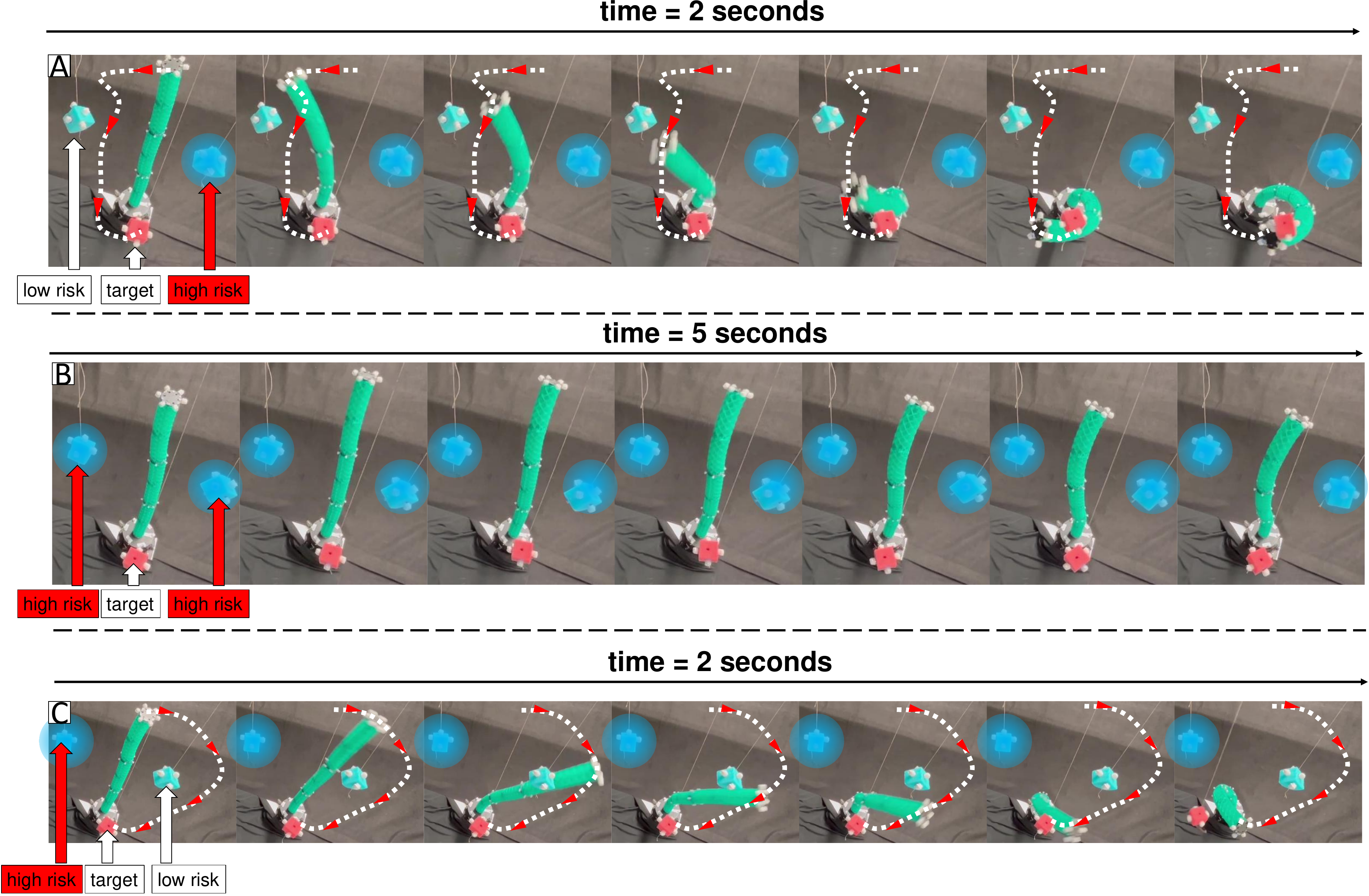}
\caption{\red{The MSRA motion for the task in obstacle avoidance control with different obstacle thresholds. Red and blue sponge cubes represent the target and obstacles. The blue transparent circles represent the high risk areas. The white dotted lines represent the MSRA end trajectories, and the red arrows represent the motion directions. Red and blue sponge cubes represent the target and obstacles.}}
\label{fig13}
\end{figure*}

\red{In real-world applications such as domestic assistance, MSRA may encounter various obstacles, including humans, taps, fragile glasses, and lit gas stoves, which lead to different safety distances.
By adjusting the obstacle threshold $r$ mentioned in Equation \ref{eq4_2}, the MSRA can show different performance on obstacles with different 'risk levels.'} 
As shown in Fig. \ref{fig13}, two obstacles are in the way of the MSRA end to the target. 
\red{These obstacles are included in the planning optimization problem with two different obstacle losses $L_{ob}$, and the respective obstacle avoidance thresholds $r$ can represent their risk levels.}
\red{If both obstacles are at the high-risk level, such as two fires on the stove, which means the obstacle threshold is large, the MSRA will vibrate near the obstacle boundaries instead of taking risks and rushing to the target, as shown in Fig. \ref{fig13}(B).
However, if one obstacle is at low risk, like a tap, S2C will automatically generate available trajectories reaching the target, and the MSRA end will move to the target, as shown in Fig. \ref{fig13}(A) and (C).}
\red{All these trajectories are planned by the optimization problem automatically, but they follow a basic rule: the MSRA end is expected to reach the target while being away from the high-risk obstacle.
For instance, in Fig. \ref{fig13}(A), our strategy first leaves away from the right high-risk obstacle and then finds a path to reach the target between two obstacles. 
In Fig. \ref{fig13}(C), our strategy first leaves away from the left high-risk obstacle and then finds a path to reach the target on the right of the right obstacles. 
With these two kinds of trajectories, our strategy achieves the task of reaching the target while avoiding the obstacles according to their risk levels.
}
All the experiment videos can be found in the Supplementary Video.

\subsection{Online Interaction}
\label{sec4.5}

\begin{figure*}[!ht]
\centering
\includegraphics[width=7.1in]{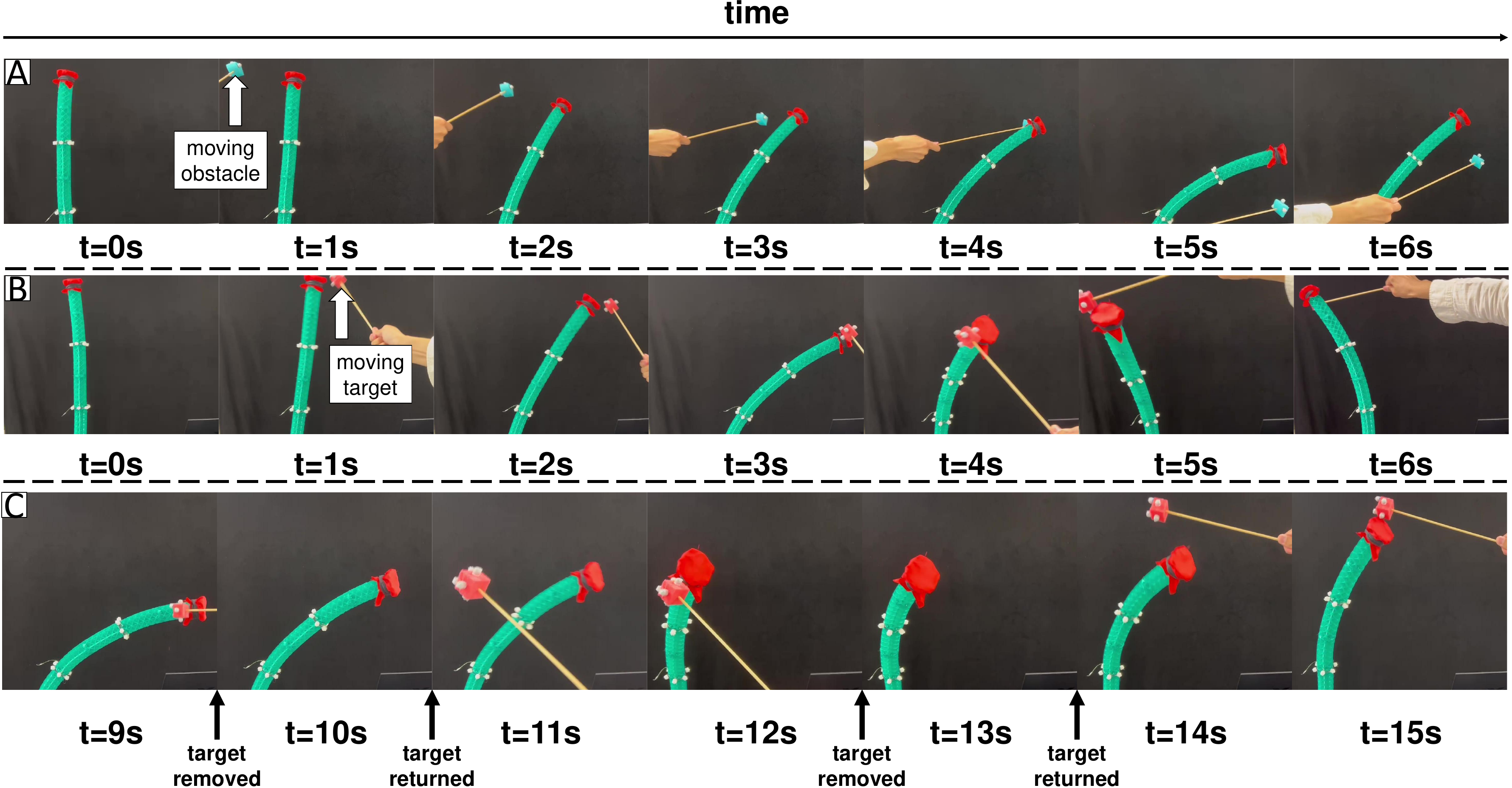}
\caption{Online planning motion for (A) obstacle avoidance and (B) target following. (C) The MSRA can reach the target even if it leaves and reenters the working space.}
\label{fig14}
\end{figure*}

If the obstacle and target will move during the task, it is necessary to plan the trajectory online. 
We include the losses $L_p, L_d, L_{ob}$ for position control and obstacle avoidance in this online task.
During the motion, we cover the optical tracker markers with a red cloth to highlight that we achieve accurate control via only motor encoder feedback. As shown in Fig. \ref{fig14}(A), a blue obstacle is moved to reach the MSRA end by hand, and the MSRA end leaves the obstacle as long as it is inside the obstacle area, whose size depends on the obstacle threshold $r$. In addition to online obstacle avoidance, target following can also be fulfilled online. In Fig. \ref{fig14}(B), the MSRA end will follow the moving target. The MSRA end will stop if the target is out of the working space, but it will start to reach the target again if the target reenters the working space, as shown in Fig. \ref{fig14}(C). All the experiment videos can be found in the Supplementary Video.

The online target following motion is steady because the MSRA always aims to minimize the distance between the target and MSRA, which is motivated by the loss $L_p$. 
Meanwhile, in online obstacle avoidance, the MSRA motion is discontinuous. When the obstacle moves near the MSRA and their distance is smaller than $r$ in Equation \ref{eq4_2}, the MSRA will start to move away from the obstacle until the distance is larger than $r$. At this time, the MSRA will stop because of the configuration change loss $L_d$ preventing unnecessary motion, which leads to such discontinuous motion. MSRA will start to move when the obstacle catches the MSRA end again. 
The obstacle threshold $r$ constraints the obstacle avoidance $L_{ob}$ into a local area, hence the loss will only take effect when the MSRA is within the 'risk area,' as the bordered blue circles in Fig. \ref{fig13} and \ref{fig14}. Therefore, we endow the MSRA with situation awareness to some extent, which may be significant in the interaction tasks in future work.

\section{Conclusion and Discussion}
\label{sec5}
This work proposes a versatile BiLSTM configuration space planning
and control strategy S2C2A specifically for MSRAs flexible to various tasks. 
We train a biLSTM network $NN_{C2S}$ mapping from configuration to task space as the MSRA forward model and leverage it in an optimization problem S2C for configuration planning. 
Then, we train a biLSTM network $NN_{C2A}$ for configuration control C2A only using inaccurate internal sensing feedback. 
Utilizing a cable-driven MSRA, we demonstrate that our approach can outperform our previous PCC strategy via position and orientation control tasks.
Furthermore, some challenging tasks have been managed, such as position constraints, obstacle avoidance, and online planning.

This flexible and versatile planning and control strategy endows MSRAs with accurate and complex deformation capability. Owing to modularity, MSRAs have the potential to perform sophisticated tasks, but effective control approaches specifically for MSRAs have not been fully explored. 
This work is a benchmark for MSRA control targeted at complicated deformation, enabling MSRA with flexibility and situation awareness. For instance, MSRAs can be utilized for minimally invasive surgery leveraging the position constraint ability and human interaction leveraging online planning ability.

There are some limitations to this strategy, which may be addressed in future work.
First, it is challenging to estimate whether one target position is feasible for one MSRA, and the difficulty will increase when including target orientation additionally.
Some physics-based approaches may be applied to MSRA working space construction.
\red{Also, this data-driven strategy may fail under an unknown payload or interaction.
An online adjustment part like the kinematics controller in \cite{23ZCb} is necessary to cope with this kind of online disturbance.
Multiple internal sensors, such as IMU and flex sensors, may contribute to the accurate configuration estimation and feedback controller.} 
Based on our data-driven strategy, we may either decrease the configuration estimation errors or apply the raw sensing feedback as a configuration without any physical models.
Moreover, we would like to take full advantage of the situation awareness ability equipped by our strategy and perform some more intricate interaction tasks.

\bibliographystyle{IEEEtran}
\bibliography{IEEEabrv,references}
\vfill

\begin{IEEEbiography}[{\includegraphics[width=1in,height=1.25in,clip,keepaspectratio]{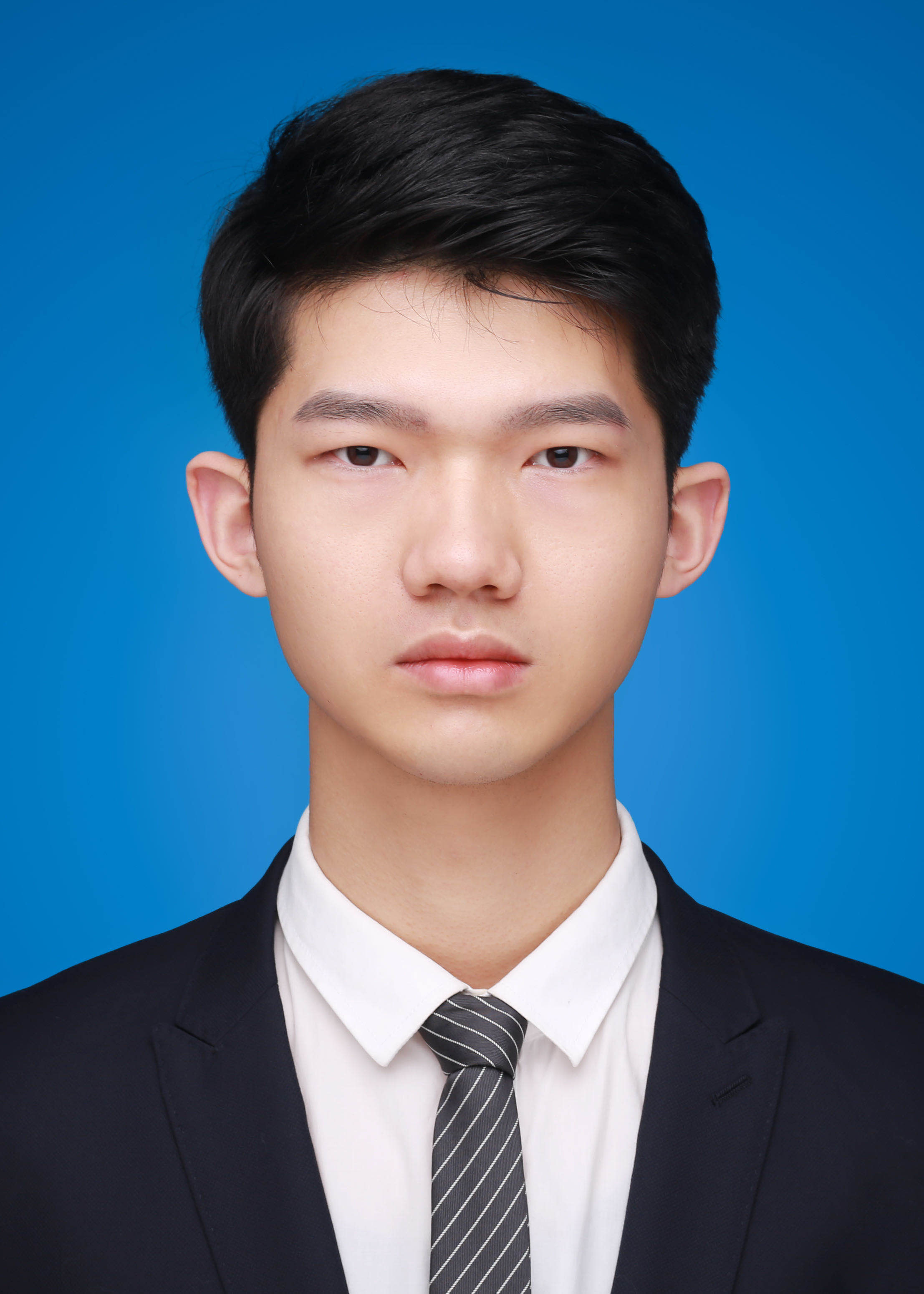}}]{Zixi Chen} received the M.Sc. degree in Control Systems from Imperial College in 2021. He is currently pursuing the Ph.D. degree in Biorobotics from Scuola Superiore Sant’Anna of Pisa.

His research interest includes optical tactile sensors and soft robot control with neural networks.
\end{IEEEbiography}

\begin{IEEEbiography}[{\includegraphics[width=1in,height=1.25in,clip,keepaspectratio]{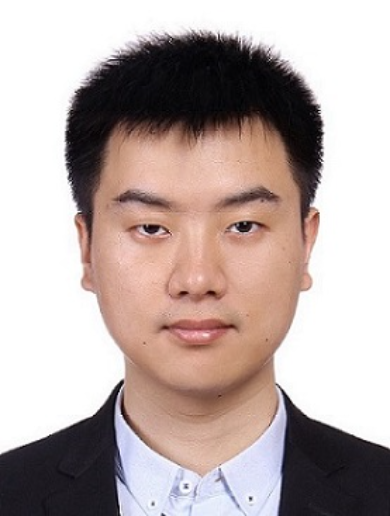}}]{Qinghua Guan} received the M.Sc. degree in 2017 and Ph.D degree in 2023. He is currently Postdoc researcher in Create Lab from EPFL. His research interest includes smart/architected materials, soft robotics and morphing structures.
\end{IEEEbiography}

\begin{IEEEbiography}[{\includegraphics[width=1in,height=1.25in,clip,keepaspectratio]{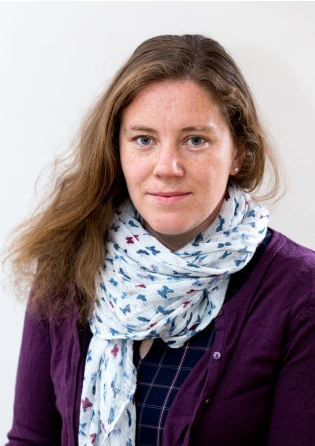}}]{Josie Hughes} is an Assistant Professor at EPFL where she established the CREATE Lab in the Institute of Mechanical Engineering in 2021. She undertook her undergraduate, masters and PhD studies at the University of Cambridge, joining the Bio-inspired Robotics Lab (BIRL). Her PhD focused on examining the role of passivity in bio-inspired manipulators, and methodologies for exploiting morphology soft large area soft sensing. Following this, she worked as a postdoctoral associate at the Computer Science and Artificial Intelligence Laboratory, Massachusetts Institute of Technology in USA in the Distributed Robotics Lab. 

Her research focuses on developing novel design paradigms for designing robot structures that exploit their physicality and interactions with the environment. This includes the development of robotic hands, soft manipulators and automation systems for applications focused on sustainability and science. Her group explore applications for agri-food, human collaboration, robot scientists and also environmental monitoring. Her work has been published in journals including Science Robotics and Nature Machine Intelligence, and she has won numerous International Robotics Competitions Awards.
\end{IEEEbiography}

\begin{IEEEbiography}[{\includegraphics[width=1in,height=1.25in,clip,keepaspectratio]{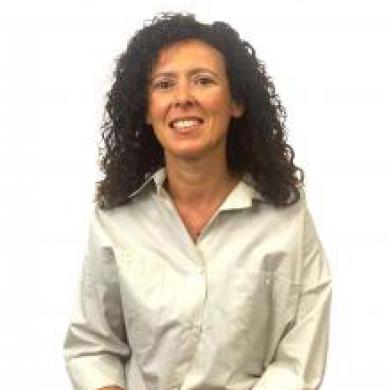}}]{Arianna Menciassi} (Fellow, IEEE) received the M.Sc. degree in physics from the University of Pisa, Pisa, Italy, in 1995, and the Ph.D. degree in bioengineering from Scuola Superiore Sant’Anna (SSSA), Pisa, Italy, in 1999.

She is currently a Professor of bioengineering and biomedical robotics with SSSA, where she is the Team Leader of the Surgical Robotics \& Allied Technologies area within The BioRobotics Institute. Since 2018, she has been the Coordinator of the Ph.D. in BioRobotics, and in 2019, she was also appointed as the Vice-Rector of the SSSA. Her research interests include surgical robotics, microrobotics for biomedical applications, biomechatronic artificial organs, and smart and soft solutions for biomedical devices. She pays special attention to the combination of traditional robotics, targeted therapy, and wireless solutions for therapy (e.g., ultrasound- and magnetic-based solutions).

Prof. Menciassi is an Editor for the IEEE TRANSACTIONS ON ROBOTICS and APL Bioengineering and an Associate Editor for Soft Robotics.
\end{IEEEbiography}

\begin{IEEEbiography}[{\includegraphics[width=1in,height=1.25in,clip,keepaspectratio]{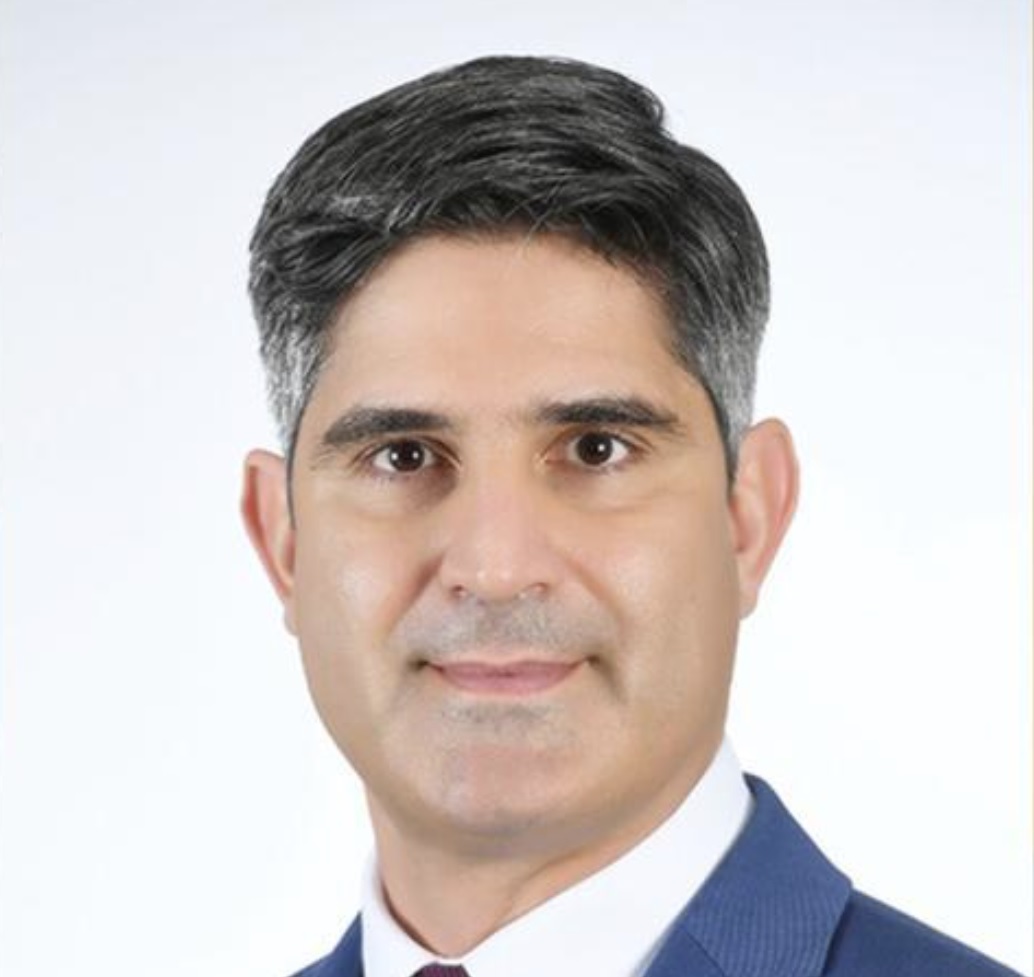}}]{Cesare Stefanini} (Member, IEEE) received the M.Sc. degree in mechanical engineering and the Ph.D. degree in microengineering, both with honors, from Scuola Superiore Sant-Anna (SSSA), Pisa, Italy, in 1997 and 2002, respectively.

He is currently Professor and Director of the BioRobotics Institute in the same University where he is also the Head of the Creative Engineering Lab. His research activity is applied to different fields, including underwater robotics, bioinspired systems, biomechatronics, and micromechatronics for medical and industrial applications. He received international recognitions for the development of novel actuators for microrobots and he has been visiting Researcher with the University of Stanford, Center for Design Research and the Director of the Healthcare Engineering Innovation Center, Khalifa University, Abu Dhabi, UAE.

Prof. Stefanini is the recipient of the “Intuitive Surgical Research Award.” He is the author or coauthor of more than 200 articles on refereed international journals and on international conferences proceedings. He is the inventor of 15 international patents, nine of which industrially exploited by international companies. He is a member of the Academy of Scientists of the UAE and of the IEEE Societies RAS (Robotics and Automation) and EMBS (Engineering in Medicine and Biology).
\end{IEEEbiography}

\end{document}